%% file: main.tex
\documentclass[sigplan,nonacm]{acmart}

\usepackage[utf8]{inputenc} 
\usepackage[T1]{fontenc}    

\usepackage{graphicx} 
\usepackage{subfigure}
\usepackage{subcaption}
\usepackage{multirow} 
\usepackage{makecell}

\usepackage{cleveref}
\begin{document}

\title{DualSparse-MoE: Coordinating Tensor/Neuron-Level Sparsity with Expert Partition and Reconstruction}

\author{Weilin Cai}
\affiliation{
  \institution{The Hong Kong University of Science and Technology (Guangzhou)}
  \city{Guangzhou}
  \country{China}
}
\email{wcai738@connect.hkust-gz.edu.cn}

\author{Le Qin}
\affiliation{
  \institution{The Hong Kong University of Science and Technology (Guangzhou)}
  \city{Guangzhou}
  \country{China}
}
\email{lqin674@connect.hkust-gz.edu.cn}

\author{Shwai He}
\affiliation{
  \institution{University of Maryland, College Park}
  \city{Maryland}
  \country{USA}
}
\email{shwaihe@umd.edu}

\author{Junwei Cui}
\affiliation{
  \institution{The Hong Kong University of Science and Technology (Guangzhou)}
  \city{Guangzhou}
  \country{China}
}
\email{jcui382@connect.hkust-gz.edu.cn}

\author{Ang Li}
\affiliation{
  \institution{University of Maryland, College Park}
  \city{Maryland}
  \country{USA}
}
\email{angliece@umd.edu}

\author{Jiayi Huang}
\authornote{Corresponding author.}
\affiliation{
  \institution{The Hong Kong University of Science and Technology (Guangzhou)}
  \city{Guangzhou}
  \country{China}
}
\email{hjy@hkust-gz.edu.cn}









\begin{abstract}
Mixture of Experts (MoE) has become a mainstream architecture for building Large Language Models (LLMs) by reducing per-token computation while enabling model scaling. It can be viewed as partitioning a large Feed-Forward Network (FFN) at the tensor level into fine-grained sub-FFNs, or \textit{experts}, and activating only a sparse subset for each input. While this sparsity improves efficiency, MoE still faces substantial challenges due to their massive computational scale and unpredictable activation patterns.

To enable efficient MoE deployment, we identify dual sparsity at the tensor and neuron levels in pre-trained MoE modules as a key factor for both accuracy and efficiency. 
Unlike prior work that increases tensor-level sparsity through finer-grained expert design during pre-training, we introduce post-training expert partitioning to induce such sparsity without retraining. 
This preserves the mathematical consistency of model transformations and enhances both efficiency and accuracy in subsequent fine-tuning and inference.
Building upon this, we propose DualSparse-MoE, an inference system that integrates dynamic tensor-level computation dropping with static neuron-level reconstruction to deliver significant efficiency gains with minimal accuracy loss.

Experimental results show that enforcing an approximate 25\% drop rate with our approach reduces average accuracy by only 0.08\%---0.28\% across three prevailing MoE models, while nearly all degrees of computation dropping consistently yield proportional computational speedups.
Furthermore, incorporating load-imbalance awareness into expert parallelism achieves a 1.41$\times$ MoE module speedup with just 0.5\% average accuracy degradation.
\end{abstract}

\maketitle

\input{Sections/Section1_Introduction}

\input{Sections/Section2_Background}
\input{Sections/Section3_Expert_Partition}

\input{Sections/Section4_Expert_Drop_System}
\input{Sections/Section5_Evaluation}
\input{Sections/Section6_Conclusion}

\bibliographystyle{ACM-Reference-Format}
\bibliography{references}










\end{document}

%% file: Sections/Section1_Introduction.tex
\section{Introduction}
\label{sec:intro}

Recently, the Mixture of Experts (MoE) architecture \cite{ShazeerMMDLHD17,gshard,rajbhandari2022deepspeed,dai2024deepseekmoe} has emerged as the mainstream design for Large Language Models (LLMs) \cite{cai2025survey,jiang2024mixtral,muennighoffolmoe,liu2024deepseekv2}, primarily due to its superior trade-off between computational efficiency and model quality. 
This trade-off is achieved through the tensor-level sparsity inherent in the MoE architecture, which can be viewed as partitioning a large Feed-Forward Network (FFN) into fine-grained sub-FFNs (termed experts) and selectively activates only a subset of these experts for processing each input token. 
By reducing computational workload compared to dense activation, existing hardware and systems can accommodate scaling model parameters to unprecedented sizes, thereby enabling higher levels of intelligence \cite{liu2024deepseek,yang2025qwen3,team2025kimi,glm45}.

Building upon the inherent tensor-level sparsity of the MoE architecture, existing research has developed specialized methods to enhance the deployment of MoE models across various scenarios. 
For instance, Expert Parallelism (EP) and its systematic optimizations \cite{rajbhandari2022deepspeed,cai2025shortcutconnected,singh2023hybrid,he2022fastermoe,liu2024deepseek} have been proposed to enable efficient distributed training and inference of MoE models; MoE compression techniques \cite{lu2024not,chen-etal-2025-eac,kim2021scalable,koishekenov2023memory}, motivated by observations of unbalanced expert selection, have been introduced to facilitate MoE model inference on edge devices. 
However, MoE models continue to present significant challenges for current machine learning systems, primarily due to their unprecedented model size and unpredictable sparse activation patterns.

To facilitate efficient post-training deployment of MoE models, we begin by analyzing the activation patterns within the MoE modules. As shown in \Cref{fig:dual-sparsity}, we observe pronounced imbalance in activation at both the tensor and neuron levels. 
We refer to this pattern as dual sparsity, wherein the output of each FFN neuron is modulated by the product of its gating score and activation value.
We further identify dual sparsity as a key factor affecting both the efficiency and accuracy of MoE models, evident in two aspects: (1) Tensor-level sparsity---prior studies \cite{dai2024deepseekmoe, he2024mixture, abnarparameters, ludziejewski2024scaling} have shown that pre-training models with increased tensor-level sparsity through more granular expert design can improve model accuracy, but may also increase gate routing overhead and lower compute intensity, leading to reduced GPU utilization; (2) Neuron-level sparsity---research on dense models has revealed a trade-off between accuracy and efficiency by omitting neurons with zero activation (as in ReLU) \cite{li2023the,liu2023deja} or negligible activation values (as in SwiGLU) \cite{zhang2024relu}.

The tensor-level sparsity configuration (expert granularity) established during pre-training may not be optimal for deployment. To address this, we propose expert partition methods (complete and partial transformations) to promote additional sparsity in the post-training phase. These methods preserve the mathematical consistency of model transformations while improving deployment efficiency.
Specifically, complete transformation converts a pre-trained MoE model into one with finer-grained experts, thereby enhancing model quality during fine-tuning.
In contrast, partial transformation targets efficiency improvements, notably by enabling our proposed Soft Expert-Tensor Parallelism (S-ETP), which provides benefits across diverse scenarios.

Moreover, we present DualSparse-MoE, an inference system that improves efficiency in a training-free manner while minimizing accuracy loss. DualSparse-MoE incorporates three key strategies:
(1) \textit{Static expert partition and reconstruction}, which divides neurons in each expert into major and minor sub-experts based on importance profiling from calibration samples;
(2) \textit{Dynamic token-expert computation dropping}, which selectively skips computations through dual thresholding of normalized gating scores, applying lower and upper thresholds to major and minor sub-experts, respectively;
(3) \textit{Load-aware thresholding}, which dynamically adjusts thresholds according to workload imbalance across devices, thereby reducing the drop rate and preserving accuracy while achieving the same speedup benefits in expert-parallel (EP) deployment.

In our experiments, we apply complete transformation to partition the Mixtral-8$\times$7B model from 8 experts into 32 finer-grained experts, reducing fine-tuning loss and improving downstream accuracy by 0.59\%.
Moreover,  partial transformation enables S-ETP, which improves EP communication efficiency in both real-world and simulated environments. 
DualSparse-MoE system demonstrates significant advantages: enforcing a approximate 25\% drop rate reduces average benchmark accuracy by only 0.08\%--0.28\% across three MoE models. 
Notably, nearly all drop rates translate directly into proportional computation reduction and speedup, a result difficult to achieve with other sparsity-based acceleration techniques. 
With load-aware thresholding in EP, our method achieves a 1.41$\times$ MoE module speedup with only 0.5\% average accuracy loss. Unlike previous selection-aware MoE compression methods designed for edge deployment, which often suffer from significant accuracy degradation and limited generalization, our approach is  tailored for distributed server-side inference, where acceleration with minimal accuracy loss is essential.

In summary, our contributions are as follows:
\begin{itemize}
    \item We identify dual sparsity in MoE architectures at the tensor and neuron levels, and demonstrate their pivotal role in balancing accuracy and efficiency.
    \item We propose expert partition methods---complete and partial transformations---to induce tensor-level sparsity during the post-training phase, achieving both accuracy and efficiency.
    \item We design DualSparse-MoE, an inference system that integrates dynamic tensor-level computation dropping with static neuron-level reconstruction, improving efficiency with minimal accuracy loss.
    \item We conduct extensive experiments to demonstrate the effectiveness of our approach in enabling more efficient post-training deployment of MoE models. 
\end{itemize}

%% file: Sections/Section2_Background.tex
\section{Background}
\label{sec:background}

\subsection{Sparsity in Mixture-of-Experts Architecture}
\label{sec:background_sparsity}

\begin{figure}[t]
    \centering
    \includegraphics[width=1\linewidth]{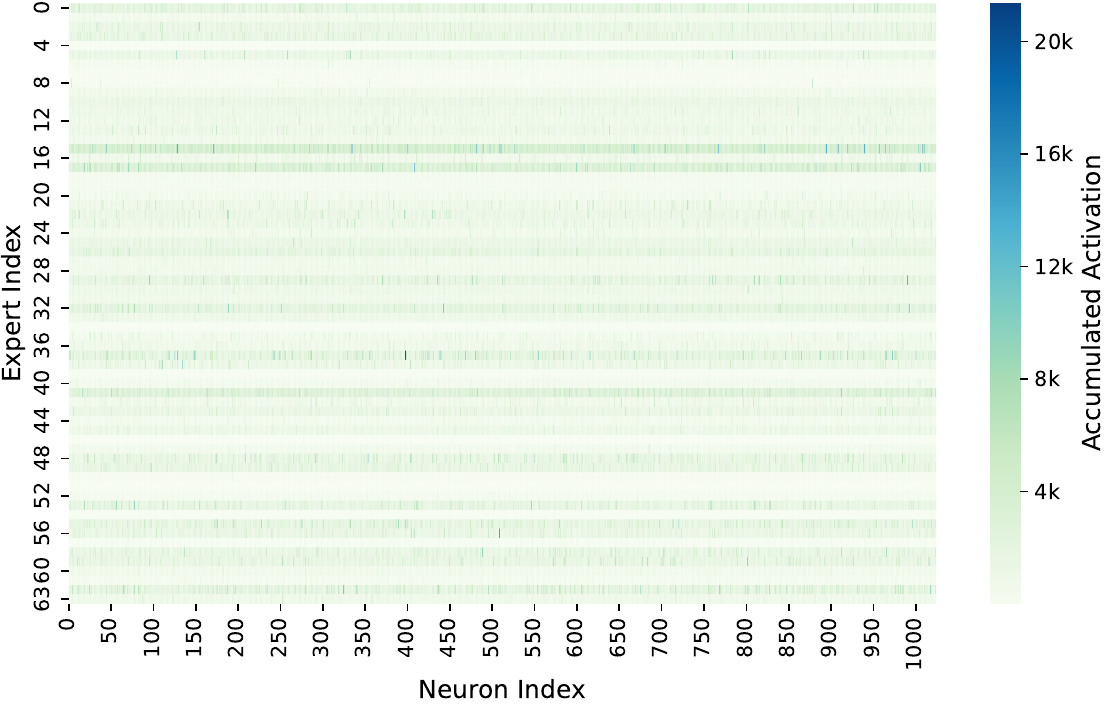}
    \vspace{-0.33in}
    \caption{Visualization of accumulated absolute activation values for each neuron across 64 SwiGLU FFN experts in a single MoE layer OLMoE \cite{muennighoffolmoe} model during inference, highlighting tensor-level sparsity (y-axis) and neuron-level sparsity (x-axis) inherent to MoE architectures.}
    \label{fig:dual-sparsity}
    \vspace{-0.12in}
\end{figure}

By visualizing MoE activation patterns during inference with the pre-trained OLMoE model \cite{muennighoffolmoe}, as shown in \Cref{fig:dual-sparsity}, we observe that the output of MoE module is governed by dual sparse at both the tensor and neuron levels. 
Specifically, color variations across rows (y-axis) reflect tensor-level sparsity, while color differences among points within each row (x-axis) capture neuron-level sparsity.

\subsubsection{Tensor-Level Sparsity}
\label{sec:background-tensor-level-sparsity}
MoE \cite{ShazeerMMDLHD17,gshard, fedus2022switch} is a neural network architecture that dynamically selects experts to process each token. An MoE layer comprises $E$ expert networks alongside a gating network $G$. 
The gating network, typically a linear network with a softmax activation function, calculates a selection probability (gating score $\mathbf{s}$) for each expert, defined as:
\begin{equation}
\label{eq:1}
\mathbf{s} = G(\mathbf{x}) = \text{Softmax}(\mathbf{x} \cdot \mathbf{W}_g),
\end{equation}
where $\mathbf{W}_g \in \mathbb{R}^{d_{model} \times E}$ is the weight matrix of the linear gating network.
Based on the gating scores, the Top-$K$ gating method is commonly used to route each input token to a subset of experts for computation, defined as:
\begin{equation}
    g_e(\mathbf{x}) = 
    \begin{cases} 
        \mathbf{s}_i & \text{if } i \in \text{TopK}(\mathbf{s}, K), \\
        0 & \text{otherwise},
    \end{cases}
\end{equation}
where $g_e(\mathbf{x})$ denotes the gating score for expert $e$. Each input token is processed by the $K$ experts with the highest gating scores, and the MoE output is a weighted sum of the outputs of the selected experts:
\begin{equation}
    \mathbf{y} = \sum_{e=1}^{E} g_e(\mathbf{x}) \cdot f_e(\mathbf{x}),
\end{equation}
where $f_e(\mathbf{x})$ denotes the output of expert $e$. Using the prevalent SwiGLU \cite{shazeer2020glu} feed-forward network (FFN) expert as an example, the expert output $f(\mathbf{x})$ is formulated as:
\begin{equation}
\label{eq:swiglu}
f(\mathbf{x}) = (\text{Swish}(\mathbf{x} \cdot \mathbf{W}_1) \odot (\mathbf{x} \cdot \mathbf{W}_3)) \cdot \mathbf{W}_2,
\end{equation}
where $\mathbf{W}_1, \mathbf{W}_3 \in \mathbb{R}^{d_{model} \times d_{ffn}}$ and $\mathbf{W}_2 \in \mathbb{R}^{d_{ffn} \times d_{model}}$ denote FFN weights, and the Swish activation function \cite{ramachandran2017searching} is used.
The three linear transformations associated with $\mathbf{W}_1$, $\mathbf{W}_3$, $\mathbf{W}_2$ are commonly referred to as the gate, up, and down projections, respectively.

Recent studies \cite{dai2024deepseekmoe, he2024mixture, abnarparameters, ludziejewski2024scaling} have systematically demonstrated that configuring experts at finer granularity---while maintaining a fixed per-token computational budget---can substantially reduce pre-training loss. 
For instance, a MoE model with 32 experts of intermediate size 1024 and Top-8 selection outperforms a MoE model with 8 experts of intermediate size 4096 and Top-2 selection. 
However, further increasing the number of experts may reduce computational efficiency due to lower compute intensity and increased gating overhead. 
These findings highlight a key trade-off: finer tensor-level sparsity can improve accuracy, but overly fine granularity may hurt efficiency.

Importantly, existing approaches determine expert granularity only before pre-training, making it difficult to adapt pre-trained MoE models to finer-grained structure for higher tensor sparsity.
To address this limitation, we propose expert partition methods that restructure experts during post-training. 
Our approach preserves mathematical consistency while improving both accuracy and efficiency in deployment.

\begin{figure}[t]
    \centering
    \includegraphics[width=1\linewidth]{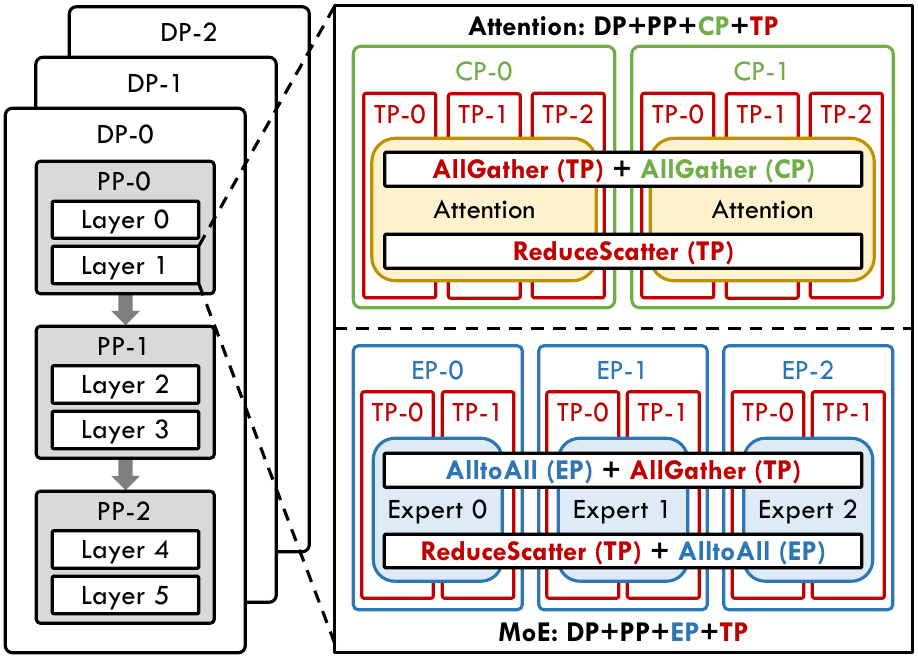}
    \vspace{-0.3in}
    \caption{An example of a state-of-the-art 5-D hybrid parallel strategy for MoE model deployment. For simplicity, the illustration omits that EP can be extended cross DP groups. Our contributions mainly focus on improving the MoE part.
    }
    \label{fig:5d-parallel}
\end{figure}

\subsubsection{Neuron-Level Sparsity}
In addition to the tensor-level sparsity inherent to the MoE architectures, prior research has investigated computation dropping and parameter pruning upon weight sparsity \cite{frantar2023sparsegpt,sunsimple,fan2025spinfer} and activation sparsity \cite{zheng2023pit,song2024powerinfer} in dense FFN. 
However, these approaches face several challenges: (1) LLMs exhibit low tolerance to high drop or prune rates, with significant accuracy degradation as these rates increase; (2) Highly fine-grained sparsity at low drop rates is difficult to translate to real speedups due to limited sparsity support in hardware and kernel designs; and (3) Most methods focus primarily on the ReLU activation function \cite{agarap2018deep}, which naturally produce zeros, and therefore cannot be directly applied to modern LLMs employing SwiGLU activations \cite{shazeer2020glu}.
In this work, we identify the activation sparsity present in MoE  FFN experts as neuron-level sparsity and propose a framework that coordinates it with tensor-level sparsity.
By jointly leveraging these two complementary forms of sparsity, we address the aforementioned challenges and improve both algorithmic accuracy and system efficiency.

\begin{figure*}
\centering
\begin{minipage}{1\linewidth}
    \subfigure[Original MoE Layer]{\label{fig:expert_partition_1}\includegraphics[width=0.29\linewidth]{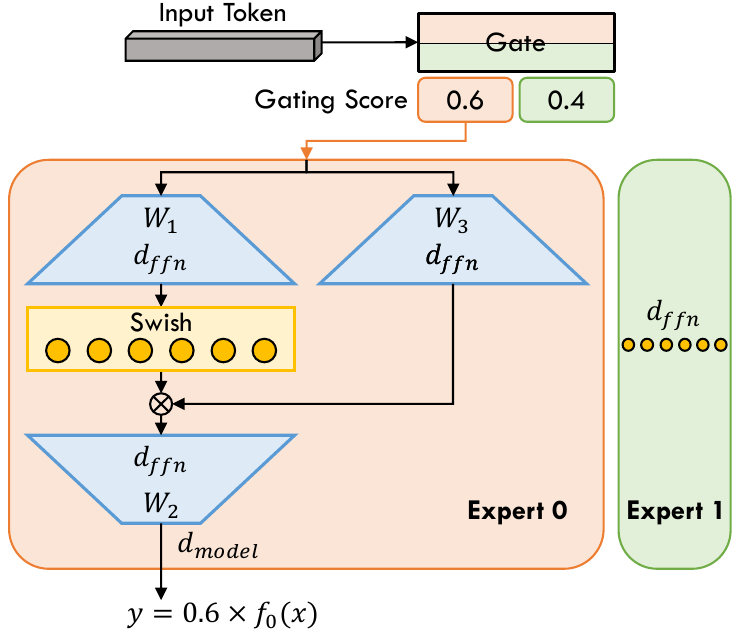}}
    \hspace{0.05in}
    \subfigure[Expert Partition (Complete Transformation)]{\label{fig:expert_partition_2}\includegraphics[width=0.345\linewidth]{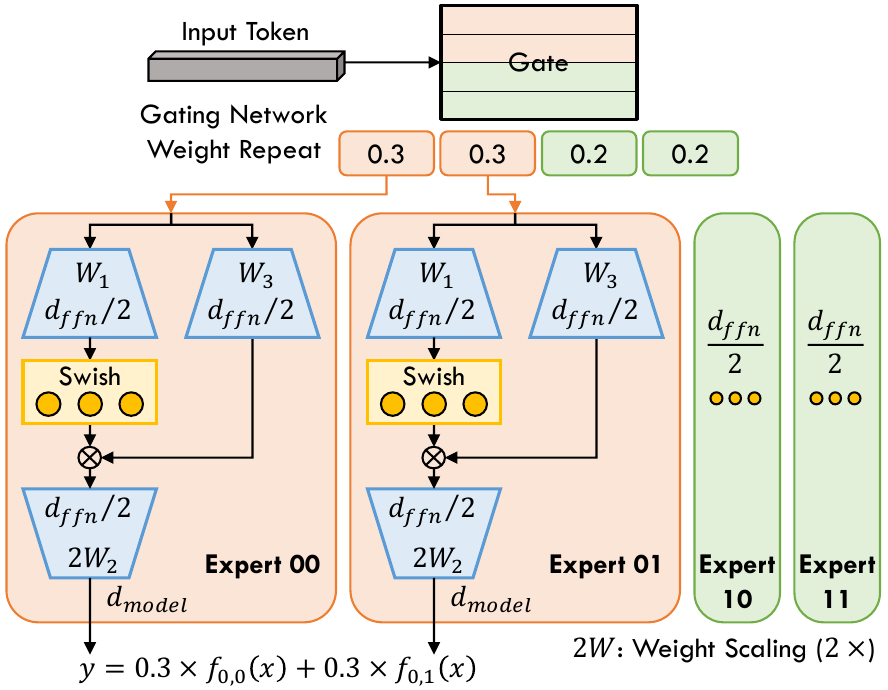}}
    \hspace{0.05in}
    \subfigure[Expert Partition (Partial Transformation)]{\label{fig:expert_partition_3}\includegraphics[width=0.345\linewidth]{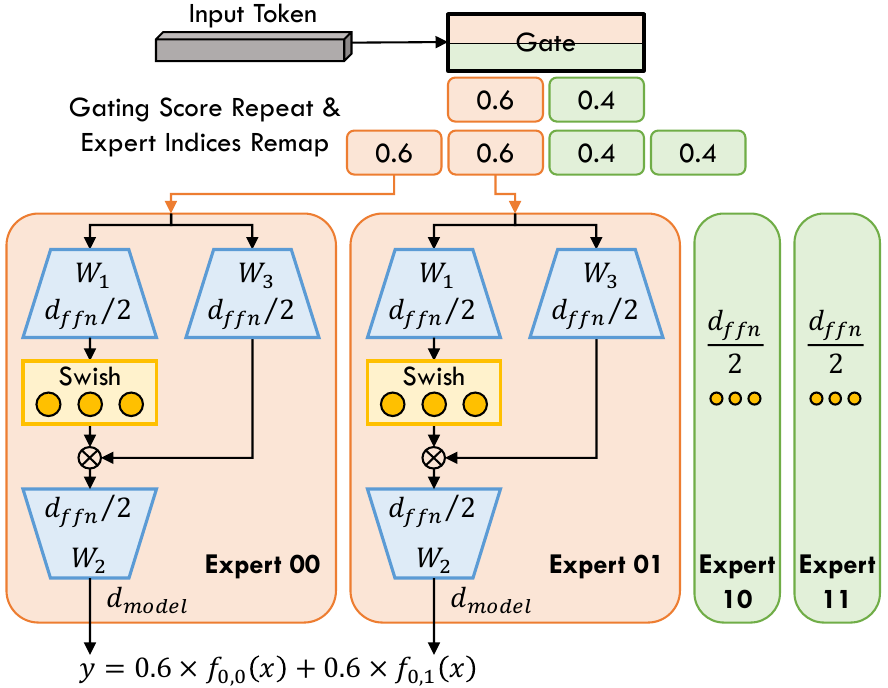}}
\end{minipage}
\vspace{-0.2in}
\caption{Illustration of expert partition methods, demonstrated by transforming a pre-trained 2-expert MoE model into a finer-grained 4-expert model. (a) Original MoE layer in the pre-trained model. (b) Complete transformation, which involves repeating the gating network weights, partitioning expert neurons, and scaling the down-projection weights $\mathbf{W}_2$. (c) Partial transformation, which involves partitioning expert neurons, repeating gating scores, and remapping expert indices.}
\label{fig:expert_partition}   
\end{figure*}

\subsection{Hybrid Parallelism for MoE Model Deployment}
\label{sec:background_parallelism}

Scaling the training and inference of MoE models across distributed devices requires an effective hybrid parallelism strategy. 
In this work, we adopt one of the state-of-the-art hybrid parallelism strategies \cite{githubGitHubNVIDIAMegatronLM}, as illustrated in \Cref{fig:5d-parallel}, to demonstrate the deployment pattern. 
This approach integrates Data Parallelism (DP) \cite{rajbhandari2020zero, ren2021zero, rajbhandari2021zero}, Pipeline Parallelism (PP) \cite{huang2019gpipe, narayanan2019pipedream}, Expert Parallelism (EP) \cite{gshard, fedus2022switch, singh2023hybrid}, Tensor Parallelism (TP) \cite{shoeybi2019megatron, smith2022using, narayanan2021efficient}, and Context Parallelism (CP) \cite{nvidiaContext_parallelPackage, korthikanti2023reducing}.
Recent studies further suggest decoupling attention and MoE modules to enable more efficient resource allocation \cite{liu2025moe, zhu2025megascale}. Although specific implementations differ---for instance, EP may be realized using either AlltoAll or AllGather---this 5-D parallelism strategy is broadly representative and offers a general reference for both training and inference.

Importantly, tensor-level sparsity directly influences the choice and configuration of parallel strategies. In MoE layers, the number of experts and their intermediate sizes primarily affect EP and TP, since these parallel strategies handle the distribution of computational and memory loads and rely on communication patterns such as ``AlltoAll+AllGather'' and ``ReduceScatter+AlltoAll.''
Therefore, the parallel strategy must comprehensively balance FLOPs utilization, communication overhead, GPU memory capacity, and other relevant factors.

%% file: Sections/Section3_Expert_Partition.tex
\section{Expert Partition}
\label{sec:expert_split}

We propose two expert partition methods: complete transformation, which restructures the model to increase tensor-level sparsity after pre-training, and partial transformation, which enables compute efficiency optimization.

\subsection{Complete Transformation}
\label{sec:expert_partition:complete_transformation}
Complete transformation partitions each expert of a pre-trained MoE model into $P$ finer-grained experts (e.g. $P=2$, as shown in \Cref{fig:expert_partition_2}).
This approach allows the transformed MoE model to integrate seamlessly with existing MoE frameworks, functioning identically to the original model.
Specifically, the transformation involves three steps:
(1) Repeat the gating network weights $P$ times and adjust the Top-$K$ selection to Top-$(K \times P)$;
(2) Evenly partition the original experts' neurons into $P$ finer-grained experts;
(3) Scale the down-projection weight $\mathbf{W}_2$ of each partitioned expert by a factor of $P$.

Next, we provide a formal derivation to demonstrate that this transformation ensures mathematical consistency.
According to \Cref{eq:1} in \Cref{sec:background_sparsity}, MoE module first employs
$\mathbf{W}_g =[h_1,h_2,\ldots,h_E]\in \mathbb{R}^{d_{model} \times E}$ to compute the gating logits $\mathbf{l}$, where each $h_i$ is a vector of dimension $d_{model}$.
Given an input token $\mathbf{x}_i$, its gating logits are computed as:
\begin{equation}
\mathbf{l} = \mathbf{x}_i \cdot \mathbf{W}_g = [l_1, l_2, \ldots, l_E].
\end{equation}
These gating logits are passed through a softmax function to obtain the gating scores $\mathbf{s}=[s_1, s_2, \ldots, s_E]$, where the gating score $s_e$ for the expert $e$ is calculated as follows:
\begin{equation}
s_e = \frac{\exp(l_e)}{\sum_{i=1}^{E} \exp(l_i)}.
\end{equation}

In the complete transformation, each vector $h_e$ in $\mathbf{W}_g$ is repeated $P$ times to construct the new gating weight matrix $\mathbf{W}_g^P \in \mathbb{R}^{d_{model} \times (E \times P)}$, defined as:
\begin{equation}
\mathbf{W}_g=[h_{1,1}, h_{1,2}, \ldots, h_{1,P}, h_{2,1}, h_{2,2}, \ldots, h_{2,P}, \ldots, h_{E,1}, h_{E,2}, \ldots, h_{E,P}].
\end{equation}
where $h_{e,p}$ denotes the $p$-th copy of the $e$-th original expert-specific vector.
Accordingly, the new gating logits $\mathbf{l}^P$ for an input token $\mathbf{x}_i$, obtained via $\mathbf{W}_g^P$, can be expressed as:
\begin{equation}
\begin{aligned}
\mathbf{l}^P &= \mathbf{x}_i \cdot \mathbf{W}_g^P 
 =[l_{1,1}, l_{1,2}, \ldots, l_{1,P}, \\ 
 &l_{2,1}, l_{2,2}, \ldots, l_{2,P}, \ldots, l_{E,1}, l_{E,2}, \ldots, l_{E,P}],
\end{aligned}
\end{equation}
where $l_{e,1}=l_{e,2}=\ldots=l_{e,P}$ due to the repeated vectors $h_{e,1}=h_{e,2}=\ldots=h_{e,P}$ in $\mathbf{W}_g^P$.
Given the extended gating logits $\mathbf{l}^P$, the gating score for each finer-grained expert $s_{e,p}$ is calculated as: 
\begin{equation}
s_{e,p} = \frac{\exp(l_{e,p})}{\sum_{j=1}^{E} \sum_{k=1}^{P} \exp(l_{j,k})}= \frac{1}{P} \cdot \frac{\exp(l_{e})}{\sum_{j=1}^{E} \exp(l_{j})}.
\end{equation}
Since all $P$ finer-grained experts partitioned from the same original expert share identical gating scores, they are activated together under the Top-$(K \times P)$ selection mechanism. Moreover, the sum of their outputs equals the original expert output, as shown below:
\vspace{-2pt}
\begin{equation}
f_{e}(\mathbf{x}_i) = \sum_{p=1}^{P} f_{e,p}(\mathbf{x}_i),
\end{equation}
which is analogous to the effect of tensor parallelism.
Consequently, the output $\mathbf{y}_i^P$ of the partitioned MoE module for an input token $\mathbf{x}_i$ can be formulated as:
\begin{equation}
\begin{aligned}
\mathbf{y}_i^P &= \sum_{e=1}^{E} \sum_{p=1}^{P} \frac{1}{P} \cdot \frac{\exp(l_{e,p})}{\sum_{j=1}^{E} \sum_{k=1}^{P} \exp(l_{j,k})} \cdot f_{e,p}(\mathbf{x}_i)  \\ 
    &= \frac{1}{P} \sum_{e=1}^{E} \cdot \frac{\exp(l_{e})}{\sum_{j=1}^{E} \exp(l_{j})} \cdot \sum_{p=1}^{P} f_{e,p}(\mathbf{x}_i) = \frac{\mathbf{y}_i}{P}.
\end{aligned}
\end{equation}
This derivation shows that the complete transformation preserves the overall MoE output by a scaling factor of $P$.

To ensure that $\mathbf{y}_i^P$ is equivalent to the original output $\mathbf{y}_i$, the result must be scaled by a factor of $P$.
There are two ways to achieve this: (1) multiplying the gating scores by $P$, or (2) scaling the expert weight $\mathbf{W}_2$ of the down-projection by $P$.
To preserve the original model structure without modifying the existing framework, we choose to scale the expert weights for complete transformation.
For example, in \Cref{fig:expert_partition_2}, the down-projection weights $\mathbf{W}_2$ of each partitioned expert is multiplied by 2, as $P=2$.

\begin{figure}
    \centering
    \includegraphics[width=1\linewidth]{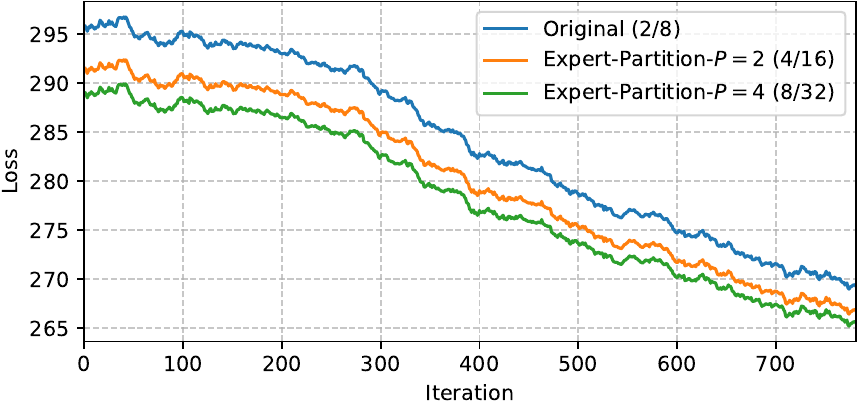}
    \vspace{-0.3in}
    \caption{Fine-tuning loss curves for Mixtral-8$\times$7B \cite{jiang2024mixtral} models under different configurations, including the original model (activating top-2 out of 8 experts) and models completely transformed with partitioned experts (activating top-4 out of 16 experts and top-8 out of 32 experts).}
    \label{fig:Mixtral_finetune_curve}
\end{figure}

\textbf{Model Quality Improvements During Fine-Tuning.}
As discussed in \Cref{sec:background-tensor-level-sparsity}, prior work has demonstrated that increasing tensor-level sparsity during pre-training---by configuring finer-grained experts---can enhance model quality.
Our proposed expert partitioning method (complete transformation) effectively promotes such sparsity for pre-trained MoE models, improving model performance during fine-tuning.
As shown in \Cref{fig:Mixtral_finetune_curve}, models with partitioned experts exhibit significantly lower loss curves compared to the original Mixtral-8$\times$7B \cite{jiang2024mixtral} model; moreover, finer-grained experts yield further loss reduction.
However, increasing the number of partitions beyond a certain point offers only marginal improvements.
Experimental results on the downstream tasks, presented in \Cref{tab:finetune-downstream} and discussed in \Cref{sec:partition-eval}, further confirm the model quality gains attributable to our approach.

\subsection{Partial Transformation}
\label{sec:expert_partition:partial_transformation}
Partial transformation offers an alternative approach for partitioning experts in pre-trained MoE models, as illustrated in \Cref{fig:expert_partition_3}.
In contrast to complete transformation, partial transformation preserves the original gating network and only modifies the results of the Top-$K$ selection through two operations: (1) repeating the gating scores and (2) remapping the expert indices.
Specifically, the original gating scores of the selected expert $[s_{1}, s_{2}, \ldots, s_{K}]$ are repeated $P$ times, resulting in $[s_{1}, s_{2}, \ldots, s_{K}]^P$. The corresponding expert indices $\mathbf{I}=[i_{1}, i_{2}, \ldots, i_{K}]$ are remapped as follows:
\begin{equation}
\begin{aligned}
\mathbf{I}^P=[i_{1}P, i_{2}P, \ldots, i_{K}P,i_{1}P+1, i_{2}P+1, \ldots, i_{K}P+1, \\ 
\ldots,i_{1}P+P-1, i_{2}P+P-1, \ldots, i_{K}P+P-1],
\end{aligned}
\end{equation}
where each original expert is partitioned and placed contiguously, maintaining its relative position.
Each expert is evenly split into $P$ finer-grained experts without scaling the down-projection weight. This is because the product of the partitioned experts' outputs and the repeated gating scores reproduces the original MoE outputs, formulated as follows:
\begin{equation}
\begin{aligned}
\mathbf{y}_i^P = \sum_{e=1}^{E} \cdot \frac{\exp(l_{e})}{\sum_{j=1}^{E} \exp(l_{j})} \cdot \sum_{p=1}^{P} f_{e,p}(\mathbf{x}_i) = \mathbf{y}_i.
\end{aligned}
\end{equation}

While partial transformation requires additional modifications to the existing MoE framework, it reduces computational overhead compared to the extended gating network required by complete transformation, despite the gating network constituting only a minor portion of the overall MoE layer’s cost.
Moreover, partial transformation maintains the original gating network parameters, enabling mathematically consistent reverse transformation and focusing solely on system efficiency. 
Consequently, we apply partial transformation to the Soft Expert-Tensor Parallelism introduced in \Cref{sec:s-etp}, as well as to the DualSparse-MoE inference system described in \Cref{sec:expert_drop:2t-drop}.

\begin{figure}
    \centering
    \vspace{-0.06in} 
    \subfigure[Expert-Tensor Parallelism (ETP)] {
        \label{fig:etp}     
        \includegraphics[width=0.98\linewidth]{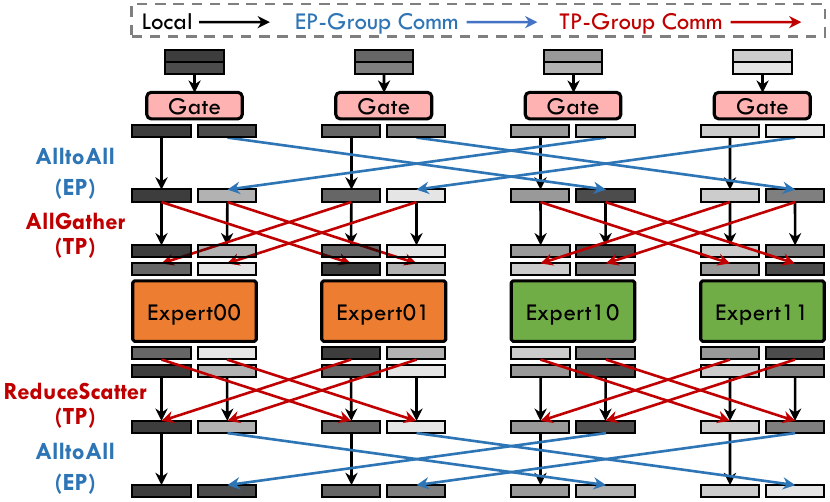}
    }
    \vskip -0.05in
    \subfigure[Soft Expert-Tensor Parallelism (S-ETP)] {
        \label{fig:s-etp}     
        \includegraphics[width=0.98\linewidth]{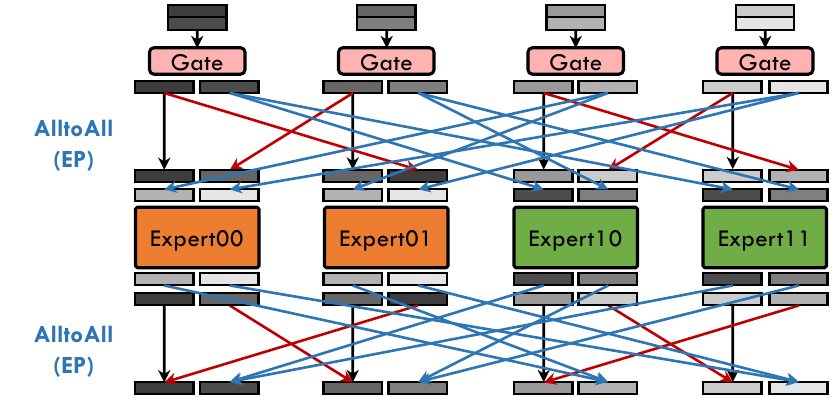}  
    }   
    \vspace{-0.2in} 
    \caption{Communication patterns in (a) Expert-Tensor Parallelism and (b) Soft Expert-Tensor Parallelism.
    }
    \vspace{-0.1in} 
    \label{fig:etp-s-etp}
\end{figure}

\subsection{Soft Expert-Tensor Parallelism (S-ETP)}
\label{sec:s-etp}
As discussed in \Cref{sec:background_parallelism}, EP and TP are employed to scale MoE deployment across distributed devices. 
In this context, applying TP to partition expert weights within EP is commonly referred to as Expert-Tensor Parallelism (ETP)~\cite{liu2025moe,singh2023hybrid}.

In contrast, we propose Soft Expert-Tensor Parallelism (S-ETP), which enables tensor-level partitioning of expert weight through an algorithmic approach rather than relying solely on system-level implementation.
Specifically, S-ETP integrates expert partition (partial transformation) with EP to achieve the same functionality as ETP.

S-ETP offers the following advantages:
\textbf{(1) Reduced Framework Complexity.} ETP often requires additional control mechanisms and framework modifications. In contrast, S-ETP addresses these challenges from an algorithmic rather than a system perspective, requiring only EP implementations, and thereby simplifying system optimization efforts.
\textbf{(2) Optimized Communication Patterns.} S-ETP uses only the AlltoAll operation (\Cref{fig:s-etp}) to achieve the same effects as the ``AlltoAll+AllGather'' and ``ReduceScatter+AlltoAll'' patterns used in ETP (\Cref{fig:etp}). This approach reduces kernel launches and synchronization overhead, improving interconnect link utilization in both training and inference.

In addition to optimizing scenarios that traditionally require ETP, our expert partition approach also benefits cases that require scaling up EP. Specifically, our method enables the deployment of a larger number of experts, thereby involving more EP devices and enhancing scalability.
Furthermore, the aforementioned advantages are also applicable to models restructured using complete expert transformation.

%% file: Sections/Section4_Expert_Drop_System.tex
\section{DualSparse-MoE Inference System}
\label{sec:expert_drop}

\begin{figure}
    \centering
    \subfigure[Expert Selection]{\includegraphics[width=1\linewidth]{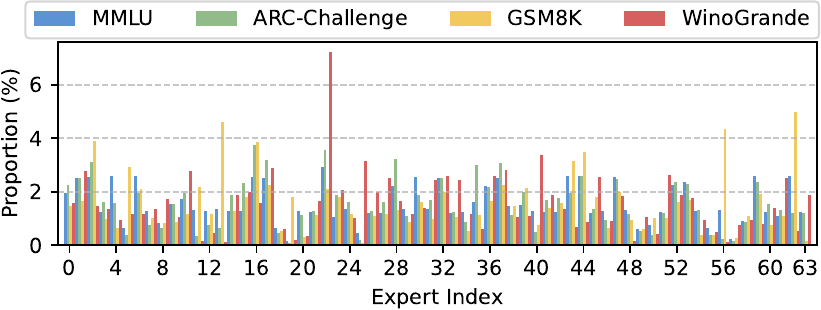}
    \label{fig:gating_analysis_a}}
    \vskip -0.06in
    \subfigure[Gating Score]{\includegraphics[width=1\linewidth]{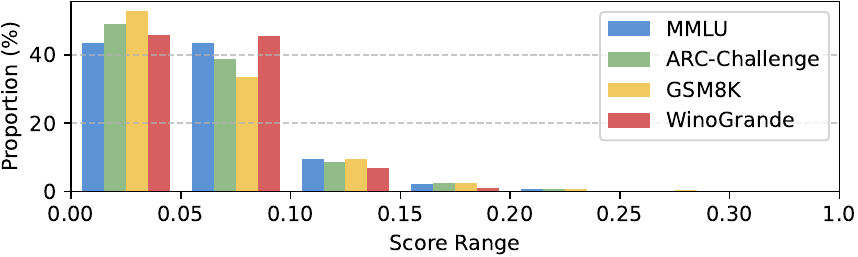}
    \label{fig:gating_analysis_b}}
    \vskip -0.06in
    \subfigure[Normalized Gating Score]{\includegraphics[width=1\linewidth]{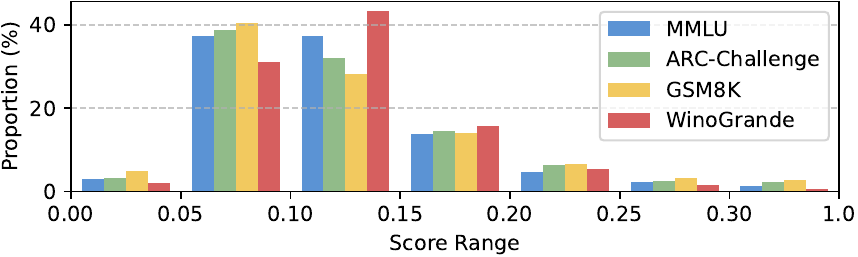}
    \label{fig:gating_analysis_c}}
    \vspace{-0.15in}
    \caption{Distributions of (a) expert selection, (b) gating scores, and (c) normalized gating scores observed during OLMoE model inference across four distinct benchmark tasks.}
    \label{fig:gating_analysis}
\end{figure}

Given the correlation between tensor-level sparsity and the superior accuracy-efficiency trade-off offered by MoE architectures, we analyze the behavior of the gating mechanism---arguably the most critical component of MoE---to identify exploitable features for inference acceleration.

Previous studies have observed imbalances in expert activation and have attempted to reduce computational cost or compress model size by skipping or pruning rarely activated experts \cite{lu2024not,chen-etal-2025-eac,kim2021scalable,koishekenov2023memory}. 
However, this approach often results in significant accuracy degradation and poor generalization across tasks, primarily due to the loss of dynamic tensor-level sparsity. 
As shown in \Cref{fig:gating_analysis_a}, expert activation patterns are highly dynamic across different benchmarks and input samples. 
We argue that diminishing the dynamic nature of tensor-level sparsity can harm model quality.

\begin{figure}
    \centering
    \includegraphics[width=1\linewidth]{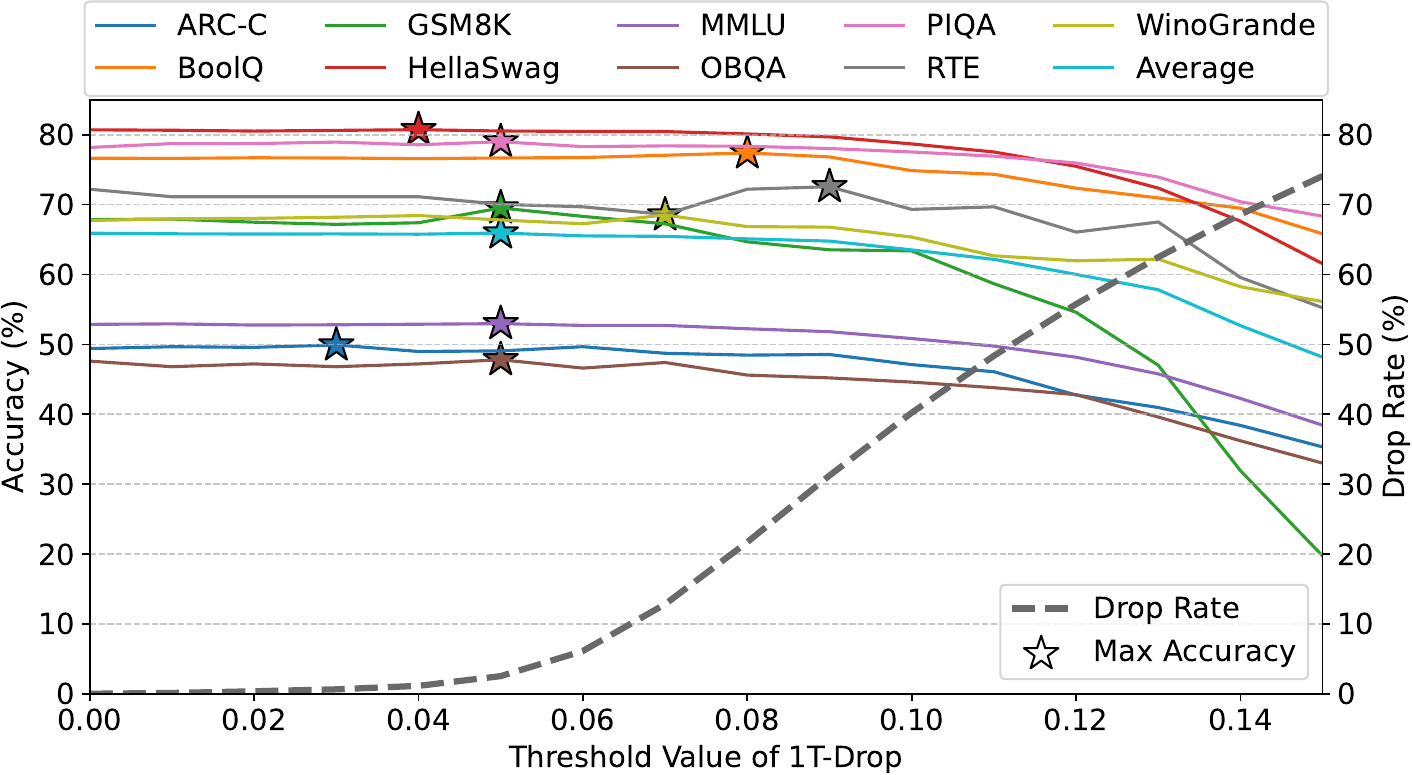}
    \caption{Benchmark accuracy and token-expert computation drop rate for OLMoE model inference using different threshold values of 1T-Drop. ``Stars'' indicate the threshold that achieves the maximum accuracy for each benchmark.}
    \label{fig:acc_drop_curve}
\end{figure}

\begin{figure*}
    \centering
    \includegraphics[width=1\linewidth]{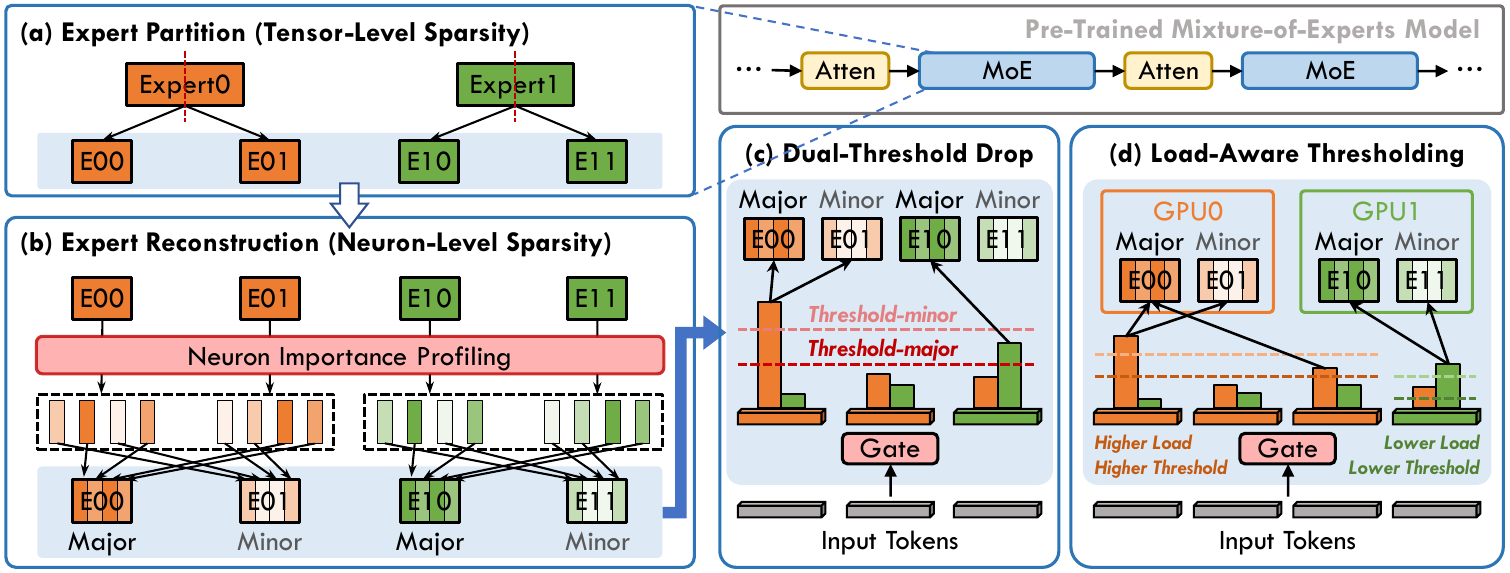}
    \caption{Overview of the proposed dual-threshold token-expert computation dropping approach (2T-Drop) and its enhancement through load-aware thresholding under EP, in the context of deploying pre-trained MoE models for inference.}
    \label{fig:2t-drop}
\end{figure*}

In contrast, our investigation reveals a relatively stable phenomenon within the gating mechanism: the distribution of gating scores for activated token-expert pairs. 
\Cref{fig:gating_analysis_b} shows that across four tasks, the distribution of gating scores is remarkably similar, with most scores falling within the ranges of 0--0.05 and 0.05--0.1, and progressively fewer scores in higher ranges.
Further normalization of the gating scores, as illustrated in \Cref{fig:gating_analysis_c}, produces a flatter distribution while preserving the consistent pattern across various tasks.

Building on this insight, we propose the DualSparse-MoE inference system, which selectively drops token-expert computations to enhance efficiency while minimizing accuracy loss. 
By exploiting the consistent characteristics of gating score distributions, our approach maintains generalizability and robust performance across a wide range of cases.

\subsection{Token-Expert Dropping via Thresholding of Normalized Gating Scores}
\label{sec:1T-drop}
Since each token-expert computation is weighted by its corresponding gating score, a lower score indicates a lesser contribution to the final result. 
In the extreme case where the gating score is zero, the computation has no effect.

To improve efficiency, we propose an operation termed ``1T-Drop'', which selectively drops token-expert computations whose normalized gating scores fall below a specified threshold ($T_{drop}^1$).
Specifically, for each input token at each MoE layer, we normalize the gating scores of the Top-$K$ activated experts and only retain experts whose normalized gating scores exceeding the threshold. The output of each token-expert computation remains weighted by its original gating score. 
It is worth noting that for some MoE models \cite{liu2024deepseek,yang2025qwen3} already normalize the gating scores of activated experts, this additional normalization step is unnecessary.

Interestingly, our empirical results show that applying a low threshold (approximately 0.05) for dropping computations can even improve accuracy, as illustrated in \Cref{fig:acc_drop_curve}. 
Across all benchmarks, the highest accuracy is achieved when some token-expert computations are dropped, suggesting that computations with very low gating scores may negatively impact overall performance. 
However, as the threshold increases further, thereby dropping more computations, the accuracy decreases across benchmarks. 
Note that the accuracy sensitivity to the drop operation also varies by task. 
For instance, GSM8K \cite{cobbe2021training}, a benchmark evaluating mathematical reasoning ability, exhibits the most pronounced accuracy decline as the drop rate increases.

\subsection{Dual-Threshold Token-Expert Dropping with Expert Partition and Reconstruction}
\label{sec:expert_drop:2t-drop}

Directly dropping token-expert computations based on a single threshold of normalized gating scores (1T-Drop) can lead to accuracy degradation, particularly at higher thresholds. Motivated by our observed dual sparsity in MoE models (\Cref{fig:dual-sparsity}), we propose a dual-threshold token-expert dropping strategy, referred to as ``2T-Drop''. It coordinates both tensor-level and neuron-level sparsity to alleviate accuracy degradation while preserving computational savings and efficiency gains.
As illustrated in \Cref{fig:2t-drop}, 2T-Drop consists of the following three key operations:

\textbf{(a) Expert Partition.} We employ the expert partition (partial transformation) method to enhance tensor-level sparsity, enabling finer-grained and thus more flexible combinations of token-expert computations dropping at the tensor level.

\begin{table*}[t]
\caption{Comparison of downstream accuracy between the original Mixtral-8$\times$7B model and its expert-partitioned variant.}
\label{tab:finetune-downstream}
\vspace{-0.1in}
\begin{center}
\resizebox{1\linewidth}{!}{
\begin{tabular}{c|c|c|c|ccccccccc|c}
\toprule
Model & $E$-Activ./Total & $T_{drop}^1$ & Drop Rate & ARC-C & BoolQ & GSM8K & HellaSwag & MMLU & OBQA & PIQA & RTE & WinoGrande & AVG.($\uparrow$)  \\
\midrule
\multirow{3}{*}{Mixtral-8$\times$7B} & 2/8 & - & 0 & 59.47 & 85.14 & 58.07 & 84.05 & 67.13 & 47.00 & 83.79 & 70.40 & 76.56 & \underline{70.18} \\
& 4/16 ($P=2$)  & - & 0 & 59.56 & 85.32 & 58.30 & 84.02 & 67.05 & 47.20 & 83.41 & 70.76 & 76.01 & \underline{70.18} \\
& 8/32 ($P=4$)& - & 0 & 59.47 & 85.26 & 58.07 & 83.99 & 67.22 & 46.80 & 83.46 & 70.76 & 76.72 & \underline{70.19} \\
\midrule
\multirow{3}{*}{\makecell{Fine-Tuned \\ Mixtral-8$\times$7B}}& 2/8 & - & 0 & 60.58 & 87.06 & 60.73 & 82.99 & 64.92 & 46.20 & \textbf{83.62} & 71.84 & \textbf{76.87} & \underline{70.53} \\
& 4/16 ($P=2$) & - & 0 & 59.56 & 87.06 & \bf 62.85 & 82.96 & \textbf{65.65} & 47.00 & 83.3 & \textbf{72.92} & 76.48 & \underline{70.86}\\
& 8/32 ($P=4$) & - & 0 & \textbf{60.67} & \textbf{87.55} & \bf 62.85 & \textbf{83.06} & 65.10 & \textbf{47.60} & 83.46 & \textbf{72.92} & \textbf{76.87} & \underline{ \bf 71.12} \\
\midrule
\multirow{3}{*}{\makecell{Fine-Tuned \\ Mixtral-8$\times$7B \\ Threshold Drop}}& 2/8 & 0.30 & 20.3\% & 59.39 & 87.06 & 61.84 & 82.72 & 64.26 & 46.40 & 82.86 & 71.48 & \textbf{76.64} & 
 \underline{70.29} \\
& 4/16 ($P=2$) & 0.15 & 21.0\% & 59.64 & 87.00 & \bf 63.46 & 82.58 & 64.75 & 46.40 & \textbf{83.13} & 73.65 & 76.24 & \underline{70.76} \\
& 8/32 ($P=4$) & 0.08 & \textbf{23.9\%} & \textbf{59.73}&\textbf{87.31}& 62.85 & \textbf{82.75} & \textbf{64.76} & \textbf{47.00} & 83.03 & \textbf{74.01} & 76.48 & \underline{ \bf 70.88}\\
\bottomrule 
\end{tabular}}
\end{center}
\end{table*}

\textbf{(b) Expert Reconstruction.}
To exploit neuron-level sparsity within each expert, we perform neuron importance profiling on calibration samples. Neurons are then reorganized to reconstruct a major sub-expert comprising neurons of higher importance and a minor sub-expert comprising those of lower importance.
Note that in our implementation, expert partitioning and reconstruction are executed as a unified process: all neurons in an original expert are first profiled and then reorganized into two separate sub-experts, one major and one minor.
We employ this static approach to leverage neuron-level sparsity and avoid the challenges of dynamically identifying neuron activations for runtime dropping.

Furthermore, we experiment with various neuron importance profiling methods within SwiGLU experts:\\ 
(1) accumulated gate value
\vspace{-1pt} 
\begin{equation}
\begin{aligned}
\text{Importance} = \sum \text{Swish}(\mathbf{x} \cdot \mathbf{W}_1^{\text{neuron}}),
\end{aligned}
\end{equation}
\vspace{-1pt}
(2) accumulated absolute gate value 
\vspace{-1pt}
\begin{equation}
 \text{Importance} = \sum \left|\text{Swish}(\mathbf{x} \cdot \mathbf{W}_1^{\text{neuron}})\right|,
\end{equation}
\vspace{-1pt}
(3) accumulated gate-up value 
\vspace{-1pt}
\begin{equation}
 \text{Importance}= \sum (\text{Swish}(\mathbf{x} \cdot \mathbf{W}_1^{\text{neuron}}) \odot (\mathbf{x} \cdot \mathbf{W}_3^{\text{neuron}})),
\end{equation}
\vspace{-1pt}
(4) accumulated absolute gate-up value
\vspace{-1pt}
\begin{equation}
 \text{Importance} \! =  \! \! \!  \sum \left|\text{Swish}(\mathbf{x} \cdot \mathbf{W}_1^{\text{neuron}}) \odot (\mathbf{x} \cdot \mathbf{W}_3^{\text{neuron}})\right|.
\end{equation}
Here, $\mathbf{W}_1^{\text{neuron}}$ and $\mathbf{W}_3^{\text{neuron}}$ denote each neuron's $\mathbf{W}_1$ and $\mathbf{W}_3$ weights, following the formulation of the SwiGLU expert in \Cref{eq:swiglu}.
Empirically, we observe that different models exhibit varying affinities for different profiling methods, highlighting the need to empirically determine the optimal configuration for each specific model.

In addition, we have considered partitioning experts into more granular based on neuron importance, which may yield higher accuracy. However, since this approach could reduce computational intensity and lead to low GPU utilization, we choose to partition and reconstruct each original expert into only two sub-experts.

\textbf{(c) Dual-Threshold Drop.}
Building on the reconstructed minor and major experts, we propose the dual-threshold drop (2T-Drop) method. This approach applies token-expert computation dropping using a higher threshold-minor ($T_{minor}^2$) for minor sub-experts and a lower threshold-major ($T_{major}^2$) for major sub-experts. Specifically, original experts with gating scores above $T_{minor}^2$ are fully engaged in computation, while those with gating scores below $T_{major}^2$ are entirely dropped---similar to the 1T-Drop method. 
Uniquely, experts with gating scores between $T_{minor}^2$ and $T_{major}^2$ compute only the major half of their neurons.
Based on our empirical experiments, we select dual thresholds of $T_{major}^2 = T_{drop}^1 - 0.01 $ and $T_{minor}^2 = T_{drop}^1 + 0.01$, which preserve a similar drop rate while achieving higher inference accuracy.

Given that our approach affects the computation granularity of token-expert grouped-GEMM and introduces additional control operations in the gating function, we optimize the corresponding Triton kernel to enhance efficiency.

\subsection{Load-Aware Thresholding in Expert Parallelism}

Load imbalance among distributed devices is a major factor limiting efficiency in MoE model inference with expert parallelism. 
Since the overall MoE computation is blocked by the device with the heaviest computational load, simply dropping computations uniformly across devices can unnecessarily degrade accuracy on devices with lighter workloads.

To address this, we propose a load-aware thresholding mechanism that dynamically adjusts token-expert dropping based on the load of each device. 
This approach enables the system to adaptively balance computation across devices while maintaining high accuracy.

As shown in \Cref{fig:2t-drop}(d), we employ a step-down thresholding strategy: devices with higher workloads apply higher thresholds, dropping more token-expert computations, while devices with lighter workloads use lower thresholds. 
To minimize control overhead in distributed environments, we calculate the ratio of the actual load to the ideal balanced load for each device. 
If this ratio exceeds 1, the threshold is set to a predefined maximum value; if it is below 1, the threshold is proportionally reduced according to the deviation from 1.
This method ensures that all devices drop computation as little as possible, while maintaining their workload at or below that of the originally most-loaded device.
By incorporating load-aware thresholding in expert parallelism, our approach achieves higher inference accuracy while maintaining the same level of acceleration.

%% file: Sections/Section5_Evaluation.tex
\section{Evaluation}
\label{sec:eval}

\begin{figure*}[t]
\centering
\begin{minipage}{1\linewidth}
    \subfigure[Real-world test conducted on a node with 8$\times$H20.] {
        \label{fig:eval_setp_real}     
        \includegraphics[width=0.47\linewidth]{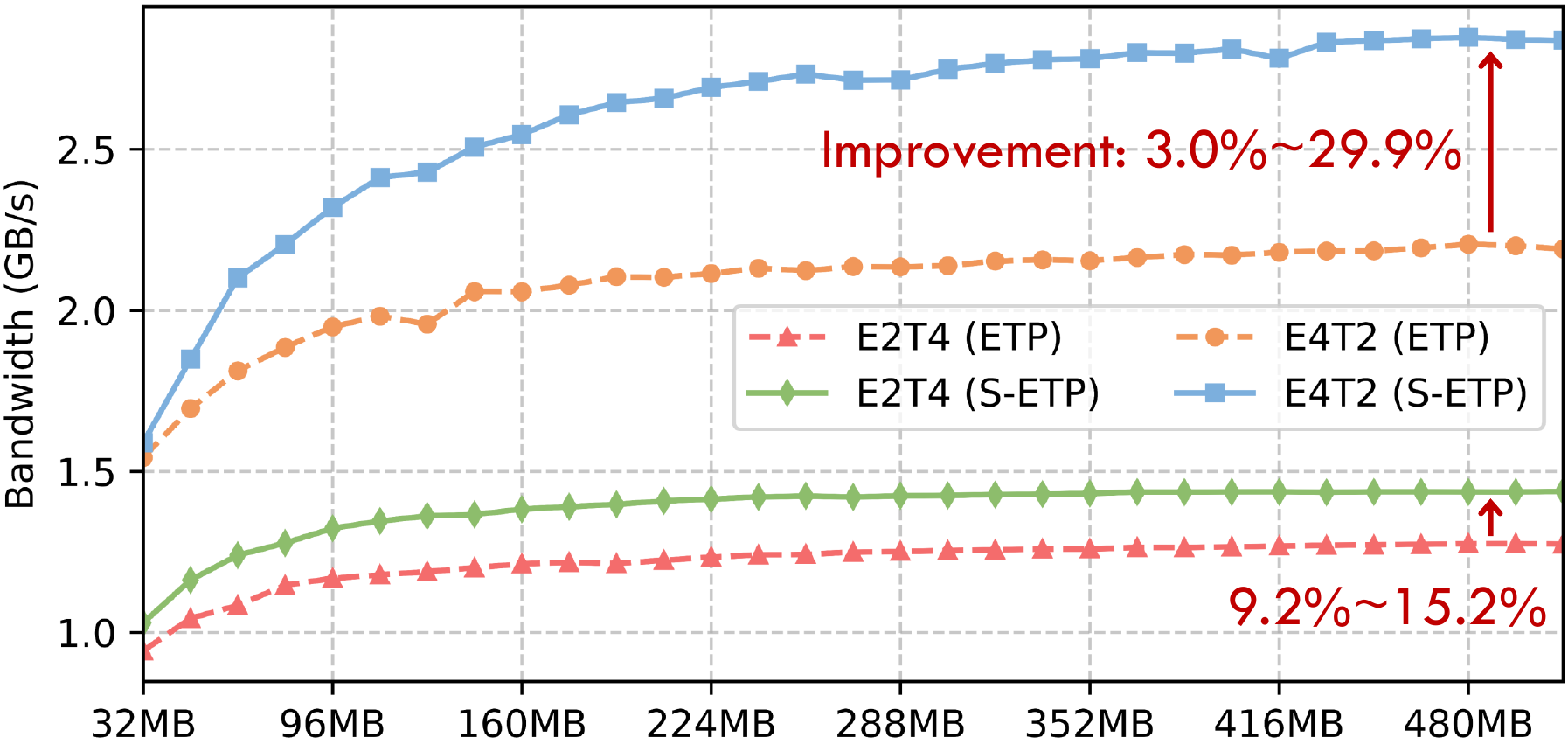}
    }
    \hspace{0.1in}
    \subfigure[Simulation performed on NVL72 and CloudMatrix384.] {
        \label{fig:eval_setp_sim}     
        \includegraphics[width=0.47\linewidth]{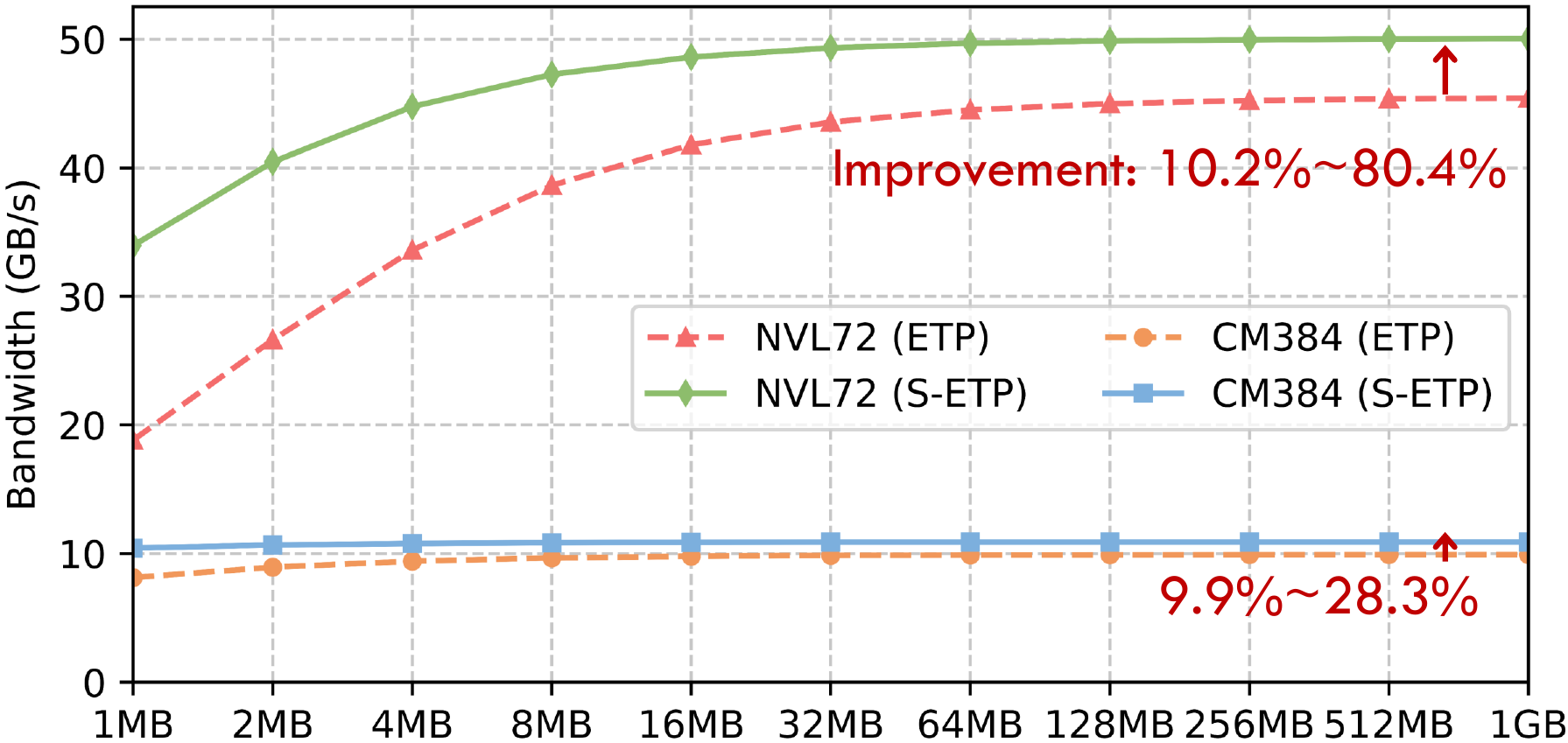}
    }   
\end{minipage}
\vspace{-0.15in}
\caption{Comparison of communication bandwidth across different input sizes using ETP and S-ETP. In real-world tests (a), ``E2T4'' denotes a configuration with EP=2 and TP=4, while ``E4T2'' denotes a configuration with EP=4 and TP=2. In simulation (b), NVL72 \cite{nvidiaNVIDIAGB200} is configured with EP=9 and TP=8, whereas CloudMatrix384 (CM384) \cite{zuo2025serving} is configured with EP=48 and TP=8.}
\label{fig:eval_setp}   
\end{figure*}

\subsection{Experimental Setup}
To evaluate the efficacy of our proposed methods, we conduct experiments on a server equipped with 8 Nvidia H20 GPUs. 
Specifically, we utilize EleutherAI's LM-Evaluation-Harness \cite{eval-harness} to assess model quality, reporting either accuracy or normalized accuracy for each benchmark, as applicable.
Our evaluation tasks include zero-shot evaluations on the ARC-C \cite{clark2018think}, BoolQ \cite{clark2019boolq}, HellaSwag \cite{zellers2019hellaswag}, MMLU \cite{hendrycksmeasuring}, OBQA \cite{OpenBookQA2018}, PIQA \cite{bisk2020piqa}, RTE \cite{wangglue}, and WinoGrande \cite{sakaguchi2021winogrande} benchmarks, as well as 5-shot evaluation on GSM8K \cite{cobbe2021training}.
We utilize the Tulu-3-sft-mixture dataset \cite{lambert2024tulu} for our fine-tuning experiments.

Furthermore, we implement our proposed DualSparse-MoE inference system and evaluate its acceleration effectiveness upon the SGLang framework \cite{zheng2024sglang}, which supports efficient distributed inference for prevailing MoE models such as Mixtral \cite{jiang2024mixtral}, OLMoE \cite{muennighoffolmoe}, and DeepSeek \cite{liu2024deepseekv2}.

Additionally, we perform small-scale real-world tests using the PyTorch Distributed framework with the NCCL backend, as well as large-scale simulations using the ASTRA-SIM simulator \cite{rashidi2020astra} to evaluate the communication optimization achieved by the Soft Expert-Tensor Parallelism (S-ETP).

\subsection{Evaluation of Expert Partition}
\label{sec:partition-eval}
We conduct experiments to substantiate the benefits of promoting tensor-level sparsity during the post-training phase, using our proposed expert partition methods.

\subsubsection{Model Quality Gains during Fine-tuning}
\label{sec:partition-eval-finetune}
We apply the expert partition (complete transformation) to the Mixtral-8$\times$7B model, partitioning its original 8 experts into 16 ($P=2$) and 32 ($P=4$) finer-grained experts.
As shown in Table~\ref{tab:finetune-downstream}, the partitioned models demonstrate the same downstream accuracy, with only negligible fluctuations. This consistency is attributed to the mathematical equivalence maintained by the partitioning process, although minor variations may arise due to floating-point precision errors. 
While models with partitioned experts exhibit significantly lower fine-tuning loss curves in Figure~\ref{fig:Mixtral_finetune_curve}, these partitioned models also achieve higher downstream accuracy after fine-tuning.
Notably, even when applying a 1T-Drop with a 23.9\% drop rate to the partitioned model ($P=4$), this model still achieves a higher average downstream accuracy of 70.88\% than the 70.53\% accuracy attained by the fine-tuned original model.

\subsubsection{Efficiency Improvements Achieved via S-ETP}
In Figure~\ref{fig:eval_setp}, our proposed S-ETP method exhibits significant improvements in communication bandwidth compared to existing ETP approach. The bandwidth is measured by dividing the input size per device by the total communication time.
In a real-world test configuration of EP=4 and TP=2 on an 8$\times$H20 node, S-ETP achieves a bandwidth improvement ranging from 3.0\% to 29.9\%. When configured with EP=2 and TP=4, the improvement ranges from 9.2\% to 15.2\%.

Furthermore, the benefits of S-ETP are particularly evident in systems equipped with fully peer-to-peer high-bandwidth interconnections, such as NVL72~\cite{nvidiaNVIDIAGB200} and CloudMatrix384~\cite{zuo2025serving}. These systems feature homogeneous network architectures, eliminating the substantial disparities typically observed between inter-node and intra-node bandwidth. 
Our simulations in these environments reveal improvements of 10.2\% to 80.4\% on NVL72 and 9.9\% to 28.3\% on CloudMatrix384.

\subsection{Evaluation of DualSparse-MoE Inference System}
Given that DualSparse-MoE inference system is proposed to enhance efficiency in a training-free manner, with minimal impact on accuracy, we evaluate it on both accuracy and efficiency perspectives in the following subsections.

\begin{table*}[t]
\vspace{-0.1in}
\caption{Comparison of downstream accuracy across different drop methods evaluated on three models. Note that setting $T_{major}^2=T_{minor}^2$ is equivalent to using 1T-Drop with $T_{drop}^1$. 2T (partition) denotes the 2T-Drop without neuron-level reconstruct.}
\label{tab:eval_drop_method}
\vspace{-0.1in}
\begin{center}
\resizebox{1\linewidth}{!}{
\begin{tabular}{c|c|c|c|c|ccccccccc|c}
\toprule
Model & Drop Method & $T_{major}^2$ & $T_{minor}^2$ & Drop Rate & ARC-C & BoolQ & GSM8K & HellaSwag & MMLU & OBQA & PIQA & RTE & WinoGrande & AVG.($\uparrow$)  \\
\midrule
\multirow{4}{*}{\makecell{Fine-Tuned \\ Mixtral-8$\times$7B \\ (8/32, $P=4$)}}& No Drop & - & - & 0 & 60.67 & 87.55 & 62.85 & 83.06 & 65.10 & 47.60 & 83.46 & 72.92 & 76.87 & \underline{71.12} \\
\cmidrule(lr){2-15}
& 1T-Drop & 0.08 & 0.08 & 23.9\% & \bf 59.73 & 87.31 & 62.85 & 82.75 & 64.76 & 47.00 & 83.03 & 74.01 & 76.48 & \underline{70.88} \\
& 2T (Partition) & 0.07 & 0.09 & 24.0\% & 59.47 & 87.25 & 63.15 & \bf 82.90 & 64.62 & 47.00 & \bf 83.51 & 74.37 & 75.85 & \underline{70.90} \\
& 2T (Reconstruct) & 0.07 & 0.09 & 24.0\% & 58.79 & \bf 87.40 & \bf 63.61 & 82.26 & \bf 64.78 & \bf 47.60 & 82.86 & \bf 74.73 & \bf 77.03 & \underline{ \bf 71.04} \\
\midrule
\multirow{4}{*}{OLMoE-Instruct}& No Drop & - & - & 0 & 49.40 & 76.64 & 67.85
 & 80.70 & 52.86 & 47.60 & 78.18 & 72.20 & 67.72 & \underline{65.91}\\
\cmidrule(lr){2-15}
& 1T-Drop & 0.08 & 0.08 & 21.7\% & 48.46 & \textbf{77.40} & 64.67 & 80.11 & 52.23 & 45.60 & 78.35 & \textbf{72.20} & \bf 66.85 & \underline{65.10} \\
& 2T (Partition) & 0.07 & 0.09 & 22.0\% & 48.12& 76.70& \textbf{68.54}& 80.04& \bf 52.81& 44.80& 77.26& 71.12& 66.61& \underline{65.11} \\
& 2T (Reconstruct) & 0.07 & 0.09 & 22.0\% & \textbf{50.00} & 77.00 & 67.22 & \bf 80.13 & 52.45 & \textbf{47.80} & \textbf{79.38} & 71.12 & 65.59 & \underline{ \bf 65.63} \\
\midrule
\multirow{4}{*}{DeepSeek-V2-Lite-Chat}& No Drop & - & - & 0 & 53.92 & 82.91 & 65.05 & 80.81 & 56.91 & 45.20 & 81.12 & 72.56 & 71.98 &  \underline{67.83}\\
\cmidrule(lr){2-15}
& 1T-Drop & 0.12 & 0.12 & 27.0\% & 51.79 & 82.91 & 63.61 & 80.30 & 55.18 & 44.40 & \bf 81.61 & 74.37 & 71.27 &  \underline{67.27} \\
& 2T (Partition) & 0.11 & 0.13 & 26.9\% & 52.13 & \bf 83.43 & 63.53 & 80.23 & 55.13 & \bf 45.80 & 80.96 &  74.01 & 71.11 &  \underline{67.37} \\
& 2T (Reconstruct) & 0.11 & 0.13 & 26.9\% & \bf 52.47 & 82.94 & \bf 64.37 & \bf 80.37 & \bf 55.58 & 44.80 & 81.39 & \bf 74.73 & \bf 72.22 &  \underline{\bf 67.65} \\
\bottomrule 
\end{tabular}}
\end{center}
\end{table*}

\subsubsection{Impact on Accuracy}
As shown in Table~\ref{tab:eval_drop_method}, applying 1T-Drop for MoE computation dropping on the evaluated models leads to a relatively significant accuracy degradation, while applying 2T-Drop with only expert partition results in a similar level of accuracy loss.
In contrast, when expert partition is combined with the reconstruction of major and minor sub-experts, 2T-Drop substantially minimizes the accuracy loss at the same drop rate. 
Specifically, imposing an approximate 25\% drop rate yields only a 0.08\% reduction in average accuracy for Mixtral, 0.28\% for OLMoE, and 0.18\% for DeepSeek.
In particular, since the DeepSeek-V2-Lite-Chat model utilizes the shared expert architecture, its drop rate is calculated as the ratio of dropped routed expert computations to the total routed and shared expert computations.

Moreover, the drop methods exceed the accuracy of the baseline with no drop in some tasks. This phenomenon may be attributed to the same factors discussed in Section~\ref{sec:1T-drop}, where applying an appropriate threshold for dropping can even enhance accuracy. Furthermore, the accuracy impact of different drop methods appears to vary across tasks.

Additionally, we conduct experiments on the fine-tuned and expert-partitioned Mixtral model, as introduced in Section~\ref{sec:partition-eval-finetune}, to demonstrate its compatibility with our proposed model transformation and inference acceleration techniques.

\begin{figure}[t]
    \centering
    \includegraphics[width=1\linewidth]{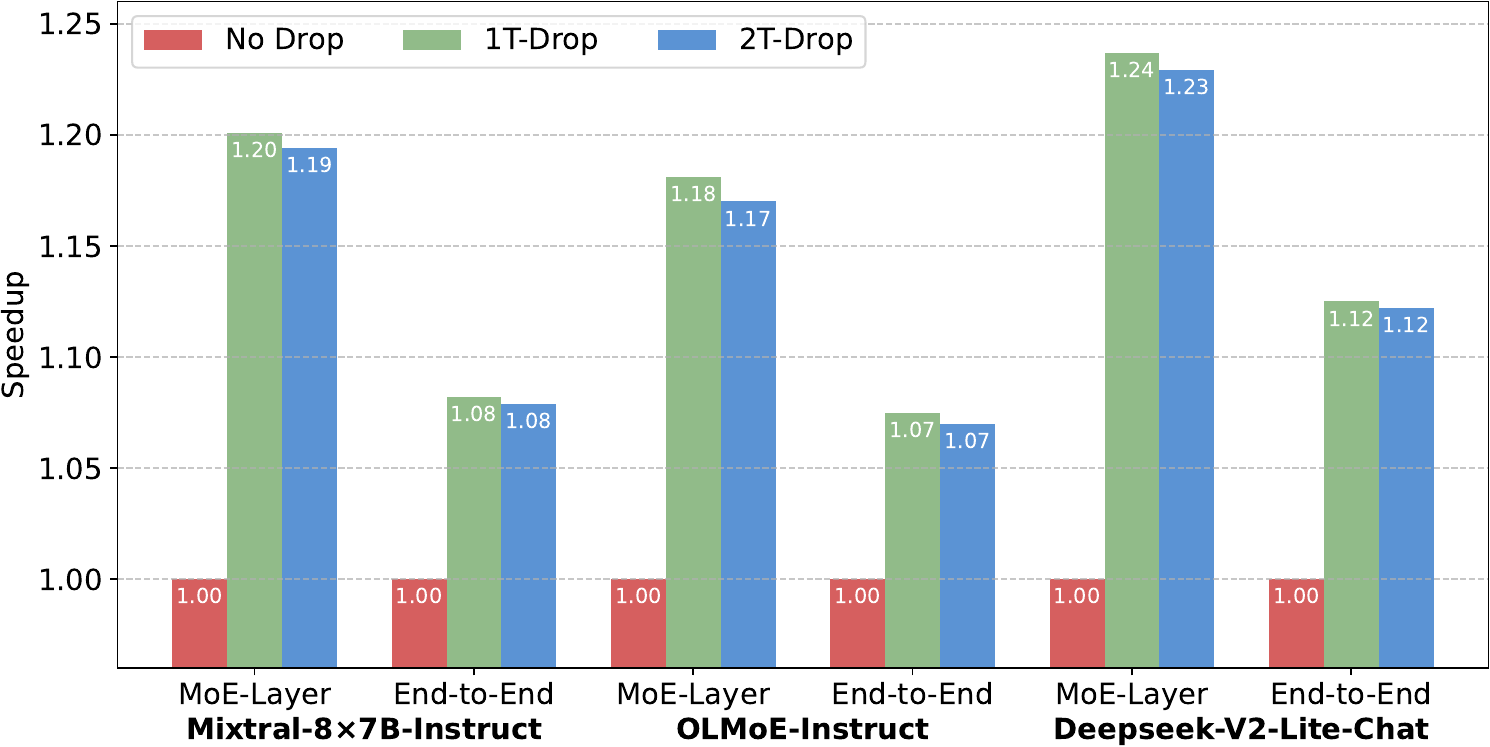}
    \caption{Comparison of the actual speedups achieved by 1T-Drop and 2T-Drop with the drop rates reported in Table~\ref{tab:eval_drop_method}. Specifically, Mixtral is deployed with TP=8 on an 8$\times$H20 node, OLMoE is deployed on a single H20 GPU, and DeepSeek is deployed with EP=8 on an 8$\times$H20 node.}
    \label{fig:eval-drop-speedup}
\end{figure}

\subsubsection{Efficiency Improvement}
Given the comparable MoE computation drop rates achieved by various drop methods, as shown in Table~\ref{tab:eval_drop_method}, we evaluate their actual speedup across different models and deployment strategies. 
To demonstrate the broad applicability of our approaches, we conduct experiments using diverse deployment strategies, including single GPU and multi-GPU setups with TP or EP.
These experiments are conducted on 2,000 random prompts with input lengths set to 500 and output lengths set to 100. The proposed drop methods are applied to achieve the drop rates shown in Table~\ref{tab:eval_drop_method}.
The results demonstrate that our methods consistently yield speedups across various deployment configurations, attributable to the reduction in computation volume achieved by our approach.

Notably, the observed MoE computation drop rates of 22\% to 27\% can be effectively translated into actual speedups for the MoE module, ranging from 1.17 to 1.23, and end-to-end speedups of 1.07 to 1.12. 
This is primarily because our methods perform dropping at the tensor level, making them well-suited for existing computing devices. 
In contrast, current sparsity-based acceleration methods struggle to achieve meaningful speedups at such low drop rates, as they require specialized hardware and custom kernels.
Furthermore, by employing optimized computing kernels, we achieve comparable speeds for 2T-Drop and 1T-Drop, even though 2T-Drop performs finer-grained computation drops than 1T-Drop.

\begin{figure}[t]
    \centering
    \includegraphics[width=1\linewidth]{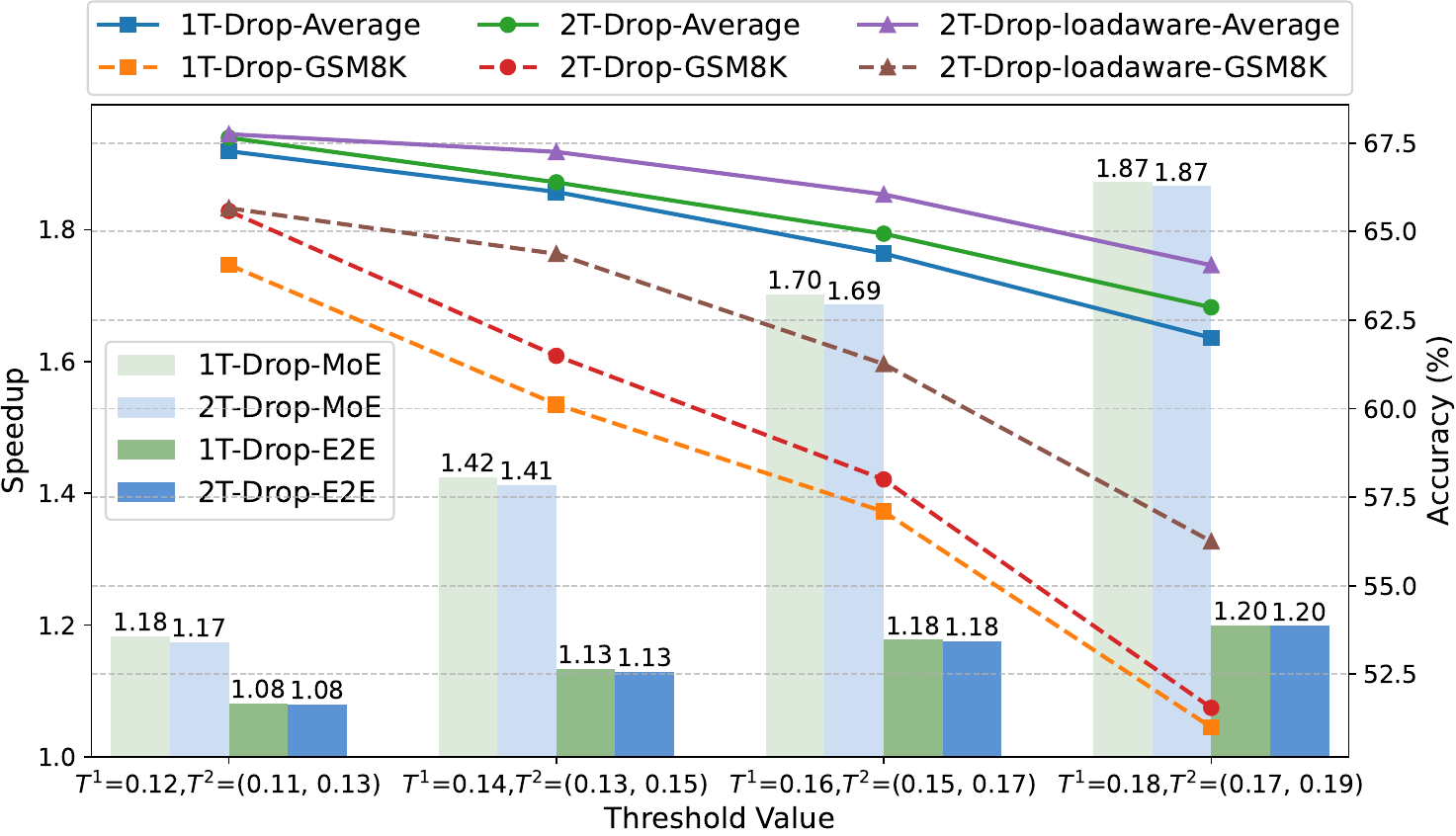}
    \caption{Comparison of speedup and accuracy among 1T-Drop, 2T-Drop, and 2T-Drop with load-aware thresholding for DeepSeek-V2-Lite-Chat model inference on an 8$\times$H20 node with EP=8. $T^1$ represents the threshold applied in 1T-Drop, while $T^2$ denotes the thresholds utilized in 2T-Drop.}
    \label{fig:eval-load-aware}
\end{figure}

\subsubsection{Improvement with Load-Aware Thresholding}
As illustrated in Figure~\ref{fig:eval-load-aware}, increasing the drop threshold results in higher acceleration but leads to a reduction in accuracy. 
Since the accuracy of the reasoning task GSM8K \cite{cobbe2021training} is particularly sensitive to the computation drop rate, we report both the accuracy on GSM8K and the average accuracy across downstream tasks presented in Table~\ref{tab:eval_drop_method}.

It is evident that 2T-Drop achieves higher accuracy compared to 1T-Drop, and that load-aware thresholding enhances the accuracy of 2T-Drop significantly. Specifically, with the integration of load-aware thresholding, 2T-Drop achieves a 1.41$\times$ speedup for the MoE module and a 1.13$\times$ end-to-end speedup, while incurring only a 0.5\% average accuracy loss.
During the practical deployment of DualSpare-MoE inference, the drop threshold can be dynamically adjusted to meet specific requirements for accuracy or throughput.

\begin{figure}[t]
    \centering
    \includegraphics[width=0.98\linewidth]{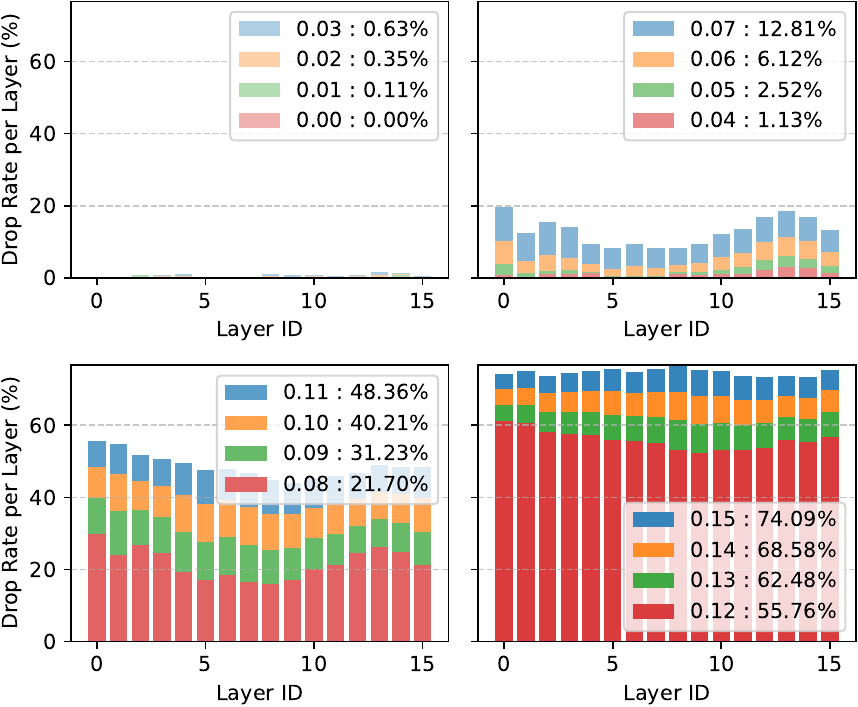}
    \caption{Drop rates across different layers of the OLMoE-Instruct model as a function of varying thresholds. The proportions in legend represent the overall drop rate of all layers.}
    \label{fig:threshold-drop-rate}
\end{figure}

Given that load-aware thresholding requires dynamic modification of the drop threshold upon device workload, we analyze the relationship between the threshold value and the computation drop rate for tested models. 
In Figure~\ref{fig:threshold-drop-rate}, which presents this relationship for the OLMoE-Instruct model, the drop rate does not change linearly with increasing threshold values. It indicates the need for a tailored mapping between threshold and drop rate.
Furthermore, the drop rate varies across different layers, suggesting potential for further exploration of per-layer thresholding strategies in future work.

\begin{figure}[t]
\centering
\begin{minipage}{1\linewidth}
    \subfigure[Gate in Low-Load Expert]{\includegraphics[width=0.496\linewidth]{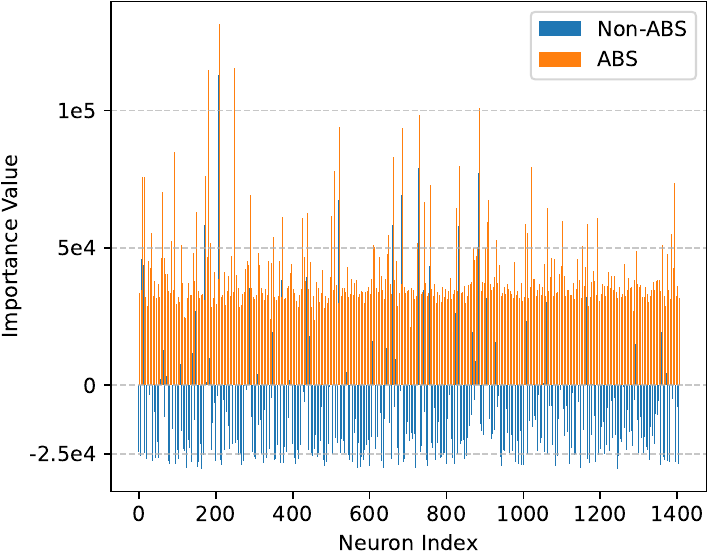}}
    \subfigure[Gate in High-Load Expert]{\includegraphics[width=0.496\linewidth]{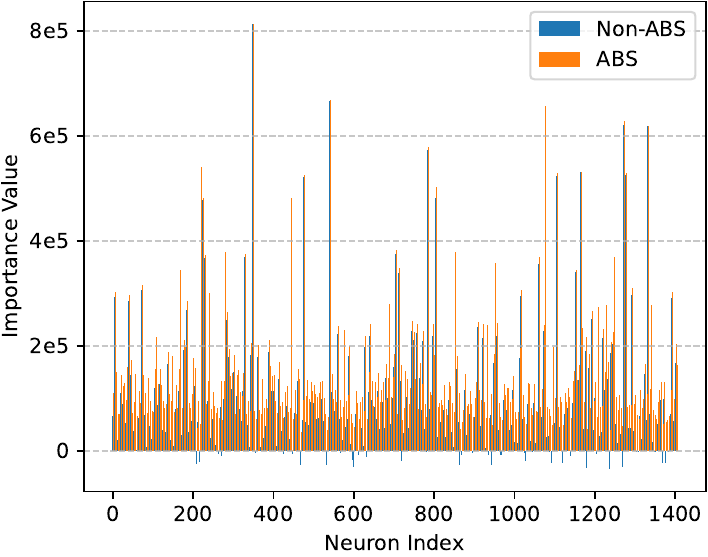}}
    \subfigure[Gate-Up in Low-Load Expert]{\includegraphics[width=0.496\linewidth]{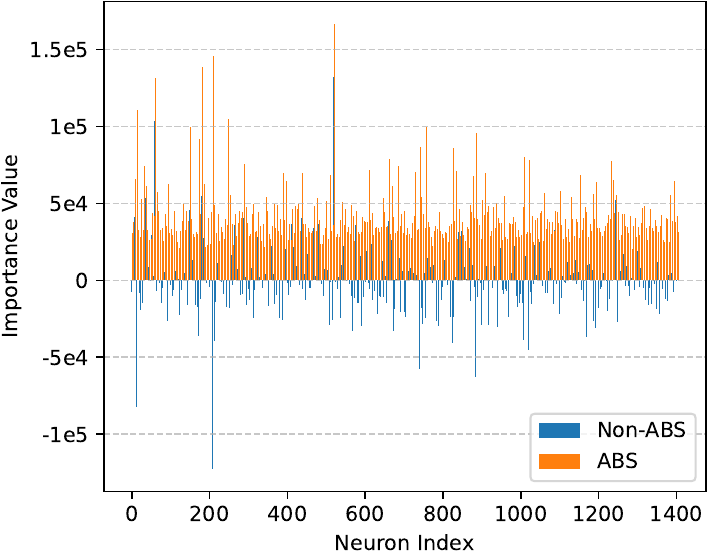}}
    \subfigure[Gate-Up in High-Load Expert]{\includegraphics[width=0.496\linewidth]{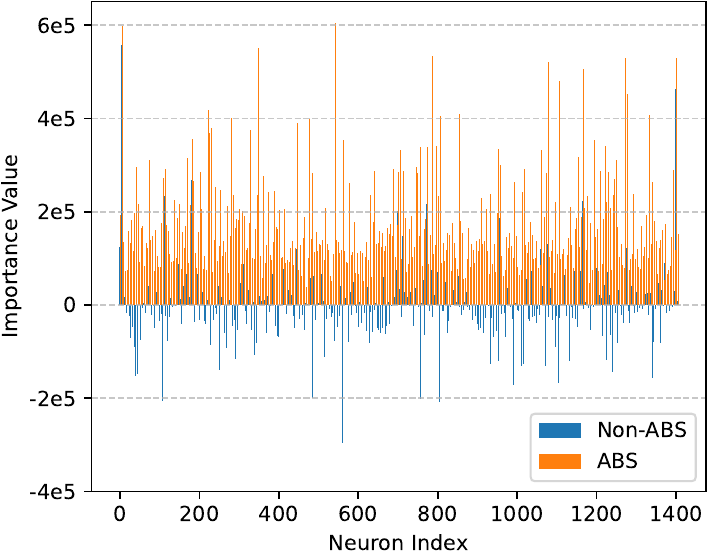}}
\end{minipage}
\caption{Comparison of neuron importance values derived from four profiling methods for expert 15 (high-load) and expert 21 (low-load) in layer 20 of DeepSeek-V2-Lite-Chat.}
\label{fig:reconstruct_analysis} 
\end{figure}

\subsubsection{Analysis of Neuron Importance Profiling}
Given the four profiling methods introduced in Section~\ref{sec:expert_drop:2t-drop}, we select the accumulated absolute gate value as the neuron importance metric for the Mixtral and OLMoE models, and the accumulated absolute gate-up value for the DeepSeek model.
Using DeepSeek-V2-Lite-Chat as an example, the average accuracy for each profiling method is as follows: 67.17 for accumulated gate, 67.29 for accumulated absolute gate, 66.79 for accumulated gate-up, and 67.65 for accumulated absolute gate-up.
These results indicate that profiling methods based on absolute values are more effective in assessing neuron importance, probably because they prevent positive and negative contributions from canceling each other out.

As shown in Figure~\ref{fig:reconstruct_analysis}, low-load experts exhibit numerous negative accumulated gate values, whereas such values are uncommon in high-load experts.
This observation suggests a potential interconnection between tensor-level and neuron-level activation, underscoring the dual sparsity characteristic of MoE architectures.
In contrast, accumulated gate-up values display similar distributions between low-load and high-load experts, which may account for the superior performance of this method in profiling the DeepSeek model.

Additionally, the selection of calibration samples also impacts the effectiveness of neuron importance profiling. 
In our experiments, we perform profiling on test models using the MMLU dataset \cite{hendrycksmeasuring}, which demonstrates strong generalization across all evaluation benchmarks.
In conclusion, neuron importance profiling methods can be further explored with different models and in various scenarios in future research.

\subsection{Advancements over Related Work}

Previous research has explored methods for exploiting sparsity to accelerate inference or achieve model compression. 
Therefore, we compare our proposed method with several prior approaches, including: Efficient Expert Skipping (EES) \cite{lu2024not}, which dynamically skips expert computation for acceleration, and Efficient Expert Pruning (EEP) \cite{lu2024not}, which permanently removes unimportant experts for compression.

EES skips the expert computation associated with the second-highest score in the Top-2 selection if the second score is less than $\beta$ times the first score, where $\beta$ is determined by the median ratio of the second score to the first score across calibration samples.
Since both EES and our proposed method are designed to reduce runtime FLOPs for inference acceleration, a fair comparison reveals that our method achieves superior accuracy (+1.2\% vs. -2.4\%) and greater speedup (1.08$\times$ vs. 1.05$\times$).

Although it is not strictly fair to compare dynamic computation reduction methods with model compression techniques, such a comparison remains informative. 
Notably, static expert pruning tends to cause substantial accuracy loss relative to dynamic computation reduction, highlighting the importance of maintaining dynamic activation within models. 
This loss is also pronounced in weight pruning methods such as Wanda \cite{sunsimple}, resulting in a substantial GSM8K accuracy reduction of 50.7\% under the 2:4 sparsity pattern.
Our proposed 2T-Drop (reconstruct) method combines dynamic tensor-level dropping with static neuron-level weight differentiation, thereby achieving enhanced performance.

Furthermore, our comparative experiments do not incorporate load-aware thresholding, as previous works have largely overlooked distributed MoE inference with EP and have not addressed its load imbalance characteristics.
It is also important to note the distinction between deployment scenarios: While edge devices prioritize model compression to accommodate limited device capacity, server-side deployments typically utilize distributed inference, where the primary focus is on maximizing accuracy and throughput rather than minimizing memory usage.

\begin{table}[t]
\caption{Comparison of our method with existing work on Mixtral-8$\times$7B-Instruct model inference. ``GSM8K Acc. Variation'' denotes the percentage change in accuracy relative to the original model. EEP \cite{lu2024not} is evaluated under two configurations, pruning 2 ($r=6$) and 4 ($r=4$) out of 8 experts.}
\label{tab:compare_related_work}
\vspace{-0.1in}
\begin{center}
\resizebox{1\linewidth}{!}{
\begin{tabular}{lccc}
\toprule
Method & Memory & Speedup & GSM8K Acc. Variation ($\uparrow$)\\
\midrule
2T-Drop (Partition) & - & $1.08\times$ & +0.5\% \\
2T-Drop (Reconstruct) & - & $1.08\times$ & +1.2\% \\
\midrule
EES & - & $1.05\times$ &  -2.4\% \\
EEP ($r=6$) & -24\% & $1.20\times$ & -8.0\% \\
EEP ($r=6$) + EES & -24\% & $1.28\times$ & -14.9\% \\ 
EEP ($r=4$) & -48\% & $1.28\times$ & -25.9\% \\ 
EEP ($r=4$) + EES & -48\% & $1.33\times$ & -36.4\% \\ 
\bottomrule
\end{tabular}
}
\end{center}
\end{table}

%% file: Sections/Section6_Conclusion.tex
\section{Conclusion}
\label{sec:conclusion}

While MoE architecture offers an excellent balance between accuracy and efficiency for LLMs through its inherent tensor-level sparsity, deploying MoE models remains challenging in current machine learning systems.
To facilitate efficient post-training deployment of MoE models, we explore the dual-sparse activation patterns present at both the tensor and neuron levels.
Specifically, we introduce expert partition to enhance tensor-level sparsity during the post-training phase, thereby improving both accuracy and efficiency.
Moreover, we present the DualSparse-MoE inference system, integrating dynamic tensor-level computation drop with static neuron-level reconstruct to accelerate inference in a training-free manner.
Experimental results show that our methods enhance efficiency while maintaining high accuracy.